\documentclass[preprint,12pt]{elsarticle}

\usepackage{amssymb}
\usepackage{amsmath}
\usepackage{booktabs}
\usepackage{graphicx}   
\usepackage{subcaption} 
\usepackage{array}      
\usepackage{todonotes}
\usepackage{algorithm}
\usepackage{algpseudocode}
\usepackage{cleveref}
\usepackage{tikz}
\usetikzlibrary{matrix} 
\usepackage{multirow}
\usepackage{threeparttable}
\usepackage{mathrsfs}
\usepackage{mathrsfs} 

\DeclareMathOperator{\kl}{KL}

\DeclareMathOperator{\argmax}{argmax}

\usepackage{float}

\journal{Signal processing}

\begin{document}

\begin{frontmatter}

\title{Scalable Context-Preserving Model-Aware Deep Clustering for Hyperspectral Images}

\author[label1]{Xianlu Li}
\author[label1,label2]{Nicolas Nadisic}
\author[label3]{Shaoguang Huang}
\author[label4,label5]{Nikos Deligiannis}
\author[label1]{Aleksandra Pi\v{z}urica}

\affiliation[label1]{organization={Ghent University}, 
            addressline={Department of Telecommunications and Information Processing}, 
            city={Ghent},
            postcode={9000}, 
            country={Belgium}}

\affiliation[label2]{organization={Royal Institute for Cultural Heritage (KIK-IRPA)}, 
            city={Brussels},
            postcode={1000}, 
            country={Belgium}}

\affiliation[label3]{organization={China University of Geosciences}, 
            addressline={School of Computer Science}, 
            city={Wuhan}, 
            postcode={430074}, 
            country={China}}

\affiliation[label4]{organization={Vrije Universiteit Brussel}, 
            addressline={Department of Electronics and Informatics}, 
            city={Brussels}, 
            postcode={1050}, 
            country={Belgium}}
\affiliation[label5]{organization={imec}, 
            addressline={Kapeldreef 75}, 
            city={Leuven}, 
            postcode={3001}, 
            country={Belgium}}

\begin{abstract}
 Subspace clustering has become widely adopted for the unsupervised analysis of hyperspectral images~(HSIs).
 Recent model-aware deep subspace clustering methods often use a two-stage framework, involving the calculation of a self-representation matrix with complexity of $\mathcal{O}(n^2)$, followed by spectral clustering. 
 However, these methods are computationally intensive, generally incorporating solely either local or non-local spatial structure constraints, and their structural constraints fall short of effectively supervising the entire clustering process. 
 We propose a scalable, context-preserving deep clustering method based on basis representation, which jointly captures local and non-local structures for efficient HSI clustering. To preserve local structure—i.e., spatial continuity within subspaces—we introduce a spatial smoothness constraint that aligns clustering predictions with their spatially filtered versions. For non-local structure—i.e., spectral continuity—we employ a mini-cluster-based scheme that refines predictions at the group level, encouraging spectrally similar pixels to belong to the same subspace. Notably, these two constraints are jointly optimized to reinforce each other. Specifically, our model is designed as an one-stage approach, in which the structural constraints are applied to the entire clustering process. The time and space complexity of our method is $\mathcal{O}(n)$, making it applicable to large-scale HSI data. Experiments on real-world datasets show that our method outperforms state-of-the-art techniques. Our experiments are reproducible, and our code is available in an online repository\footnote{https://github.com/lxlscut/SCDSC}.
\end{abstract}

\begin{graphicalabstract}
\begin{figure}[ht]
    \centering
    \includegraphics[width=\linewidth]{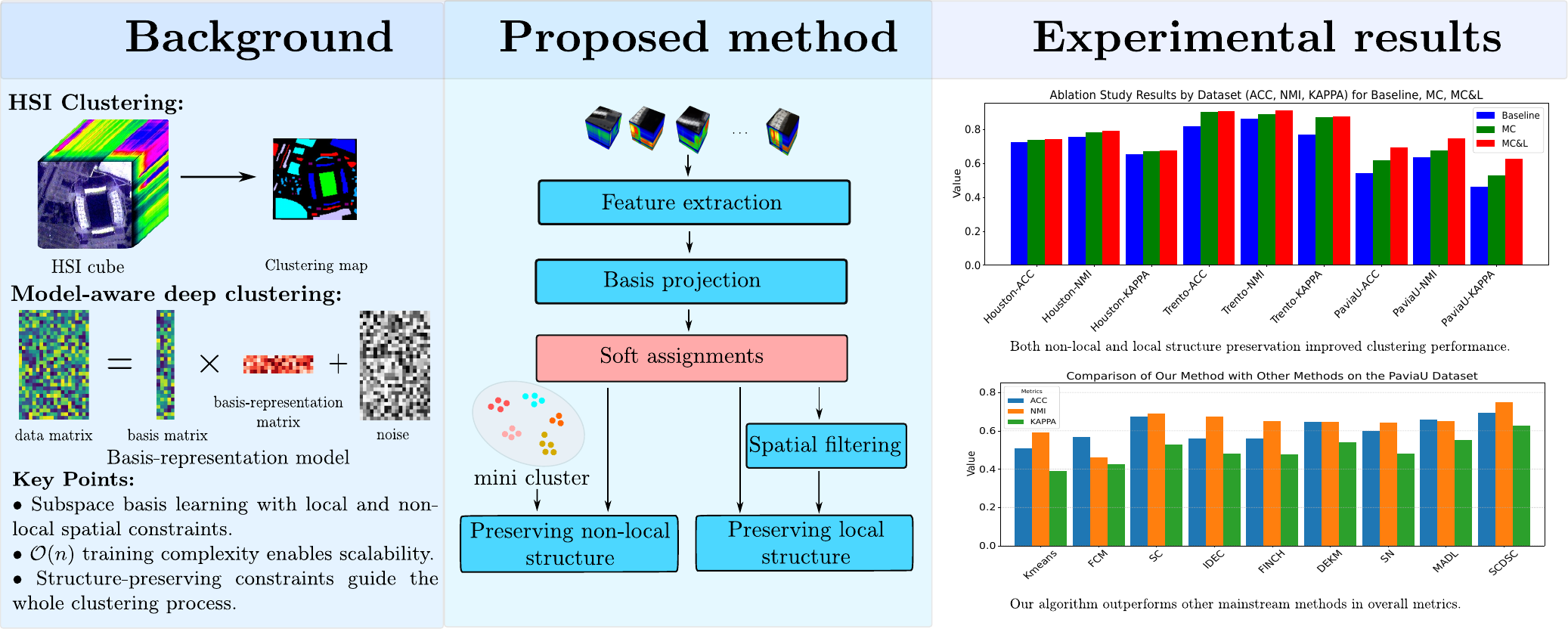}
        \captionsetup{labelformat=empty}
    \caption{Graphic abstract}
\end{figure}
\end{graphicalabstract}

\begin{highlights}
\item We propose an end-to-end clustering network to group HSI pixels by learning the bases of different subspaces, guided by both local and non-local spatial structures.
\item We propose a mini-cluster updating scheme that first groups pixels with their nearest neighbors in the spectral space to form mini-clusters. It then updates the soft assignment at the mini-cluster level, encouraging shared predictions and preserving the non-local structure. This method operates with a complexity of $\mathcal{O}(n)$, offering efficiency gains over Laplacian regularization.


\item We propose a local spatial structure constraint that preserves spatial continuity in HSI data by promoting alignment between original prediction and smoothed predictions generated from spatial filtering. Unlike previous works where the spatial constraint is applied to the self-representation matrix with a complexity of $\mathcal{O}(n^2)$, our method operates with a reduced complexity of $\mathcal{O}(n)$. More importantly, our constraint provides end-to-end supervision in the entire training process, offering stronger guidance for model optimization.

\item We evaluate the proposed method on three benchmark HSI datasets, demonstrating that our method outperforms other mainstream approaches. An ablation study further confirms the effectiveness of both the local and non-local structure preservation components.
\end{highlights}


\end{frontmatter}

\section{Introduction}
\setcounter{figure}{0}
Hyperspectral images~(HSIs) record the precise electromagnetic spectrum of the objects in a scene in hundreds of spectral bands, enabling this way discrimination between objects that are indistinguishable in conventional RGB images. As a result, HSIs have been widely applied in fields such as agriculture~\cite{liu2017hyperspectral}, environmental monitoring~\cite{stuart2019hyperspectral}, and defense and security~\cite{briottet2006military}.
Clustering, which aims to categorize image pixels into different classes without any labeled data, plays a crucial role in interpreting HSI data. However, HSI clustering remains challenging due to the presence of noise and high spectral variability~\cite{huang2023model}. Subspace clustering~\cite{SSC, vidal2011subspace}, which models data as lying in a union of low-dimensional subspaces, has shown strong performance in this context and has gained significant attention in recent years.
Subspace clustering assumes that high-dimensional data lies in a few low-dimensional subspaces, where the data points within the same subspace are seen as a class. Representative subspace clustering methods can be categorized into model-based subspace clustering \cite{SSC,liu2012robust,wang2013provable,tian20200}, and model-aware deep subspace clustering \cite{DSCNet,peng2020deep,9440402}. The model-based subspace clustering methods include sparse subspace clustering~(SSC)~\cite{SSC}, low-rank representation~(LRR)~\cite{liu2012robust} and joint SSC~(JSSC)~\cite{huang2018joint}. 
These approaches are designed based on the self-representation property, which assumes that each data point in a subspace 
$\mathcal{S}_i$ can be expressed as a linear combination of other points within the same subspace, subject to sparsity or low-rank constraints.
The resulting representation matrix effectively reveals the affinities between different data points. It is thus used to construct a similarity matrix, which is further fed to spectral clustering to obtain the clustering result. However, these methods are limited by the matrix-decomposition-based shallow representations, making it difficult to cluster HSIs that are often non-linearly separable in practice. 

To solve this problem, model-aware deep subspace clustering methods leverage the feature extraction capacity of deep neural networks to extract discriminative features and take into account non-linear interactions.
Representative methods include DSCNet~\cite{DSCNet}, DSC~\cite{peng2020deep}, SDSC-AI~\cite{li2021self}, LRDSC~\cite{zeng2019spectral} and PSSC~\cite{9440402}. 
Typically, autoencoders are used to project the input data onto a latent feature space, then a fully connected layer is incorporated within the latent space, positioned between the encoder and decoder, to approximate the self-representation model.
Li et al.~\cite{li2021self} use clustering results as pseudo labels to train the feature extraction network, enhancing feature discriminability for clustering tasks. They also initialize the self-representation layer with a K-Nearest Neighbors~(KNN) graph to reduce dictionary redundancy, leading to significant performance improvements. In \cite{valanarasu2021overcomplete}, features from undercomplete and overcomplete autoencoders are fused for subspace clustering, achieving outstanding performance without pre-training. Chen et al.~\cite{chen2021novel} propose leveraging self-attention within the autoencoder to capture long-range dependencies, yielding better results than DSCNet. Benefiting from the improved feature representation in the latent space with the encoders, model-aware deep subspace clustering methods are more effective in handling data of complex structures compared with the aforementioned model-based subspace clustering methods. By optimizing the parameters of the fully connected layer, the self-representation matrix can be obtained for the construction of the similarity matrix. However, they still suffer from the following issues. 
First, these methods are computationally expensive. This is because the self-representation matrix is of size $n \times n$ (where $n$ is the number of HSI pixels), leading to a training complexity of $\mathcal{O}(n^2)$ that makes large-scale clustering impractical; in addition, the spatial constraints employed further increase the computational burden. Second, the features extracted may not be optimal for clustering. This is due to that feature extraction and spectral clustering are performed separately, which risks degrading overall performance. Third, these methods struggle to capture the intrinsic cluster structure of HSIs. This limitation arises from their focus on either local or non-local spatial dependencies, rather than fully exploiting the spatial relationships present in the data.

In this paper, we propose a scalable context-aware deep subspace clustering~(SCDSC) method, which performs feature extraction and clustering within a unified framework. In contrast to conventional self-representation-based clustering methods that require optimizing a large self-representation matrix, we follow the approach proposed in~\cite{EDESC} and instead learn compact subspace bases.
Those bases are compact, class-specific subspace dictionaries with fewer parameters in the latent feature space.
We then obtain directly the clustering soft assignment by projecting the latent representation onto the subspace bases. 
The resulting model has a low computational complexity, supporting scalability and efficient processing of large HSIs.

To capture both local and non-local dependencies in hyperspectral data, we introduce two structural constraints. The local structure constraint enhances spatial homogeneity in the clustering results and improves robustness to noise and spectral variability. We do so using a spatial-wise mean filter to smooth the clustering results. The non-local structure constraint promotes consistency among spectrally similar data points, regardless of their spatial distance, by grouping them into mini-clusters and encouraging shared cluster assignments within each group. In contrast to existing local spatial constraints like total variation, which exhibit high computational complexity, our method is computationally efficient. The improved local homogeneity can be propagated to non-local data points through the non-local constraint, and conversely, the non-local constraint can also enhance local homogeneity, creating a mutually reinforcing relationship. To the best of our knowledge, this is the first attempt to develop an end-to-end, scalable deep subspace clustering method for HSIs.
Experimental results on three benchmark datasets show that our method consistently outperforms several state-of-the-art methods, both model-based and deep-learning-based. A preliminary version of this work was presented in~\cite{li2023model}, where we applied spatial filtering to embed spatial continuity into the soft assignment optimization. 
To preserve the non-local structure, \cite{li2023model} uses contrastive learning in the feature space. Compared with that preliminary work, we develop a novel approach to modeling non-local similarities by means of a mini-cluster grouping. 
Moreover, we provide a more detailed presentation, a deeper analysis of the overall approach, and critical discussions.
We also present a more extensive experimental study.

The remainder of the paper is as follows: 
Section 2 provides a comprehensive analysis of model-based subspace clustering and deep clustering methods, including purely data-driven and model-aware approaches. 
Section 3 describes our main contribution, a context-aware deep subspace clustering method. 
Section 4 evaluates it using three real-world hyperspectral datasets. 
Finally, Section 5 concludes this paper.

\section{Related work}
In this section, we introduce the key concepts and models that form the foundation of our proposed method.
Then, we briefly review the existing approaches for HSI clustering.

\subsection{Agglomerative hierarchical clustering} 
\label{subsec:finch}
    In our work, we use agglomerative hierarchical clustering to generate mini-clusters, as will be detailed in Section 3. Agglomerative hierarchical clustering is a method that groups data points by gradually merging clusters. It starts with each data point as its separate cluster, and then at each iteration, it merges clusters based on a defined rule, called a linkage criterion. As a variation of this method, FINCH~\cite{finch} uses the first neighbor of each sample to identify long neighbor chains and uncover groups within the data. This method shows high performance in clustering complex data with a complexity of $\mathcal{O}(n \log n)$ where n is the number of data points. In every iteration of FINCH, the first nearest neighbor adjacency matrix is created as follows:
    \begin{equation}
        \mathbf{A}(i,j) = 
        \begin{cases} 
        1 & \text{if } j = \kappa^{1}_{i} \text{ or } \kappa^{1}_{j} = i \text{ or } \kappa^{1}_{i} = \kappa^{1}_{j} \\
        0 & \text{otherwise} 
        \end{cases}
    \end{equation}
    where $\kappa^{1}_{i}$ denotes the first nearest neighbor of point $i$. During the merging process, data points connected by the same neighbor chain are grouped, and a new data point is generated as the mean of these points. 
    The generated data points will be merged in the same way in the next iterations.
    In the initial few iterations, the relationships between data points are simpler, allowing the algorithm to merge data points accurately. 
    Compared to methods that require a predefined number of neighbors, FINCH is a parameter-free algorithm, and it allows clusters of different sizes.
    This results in a more flexible and adaptive capture of non-local data structures.


\subsection{Model-based subspace clustering}
\begin{figure}[!t]
                    \centering
                    \includegraphics[width=0.7\linewidth]{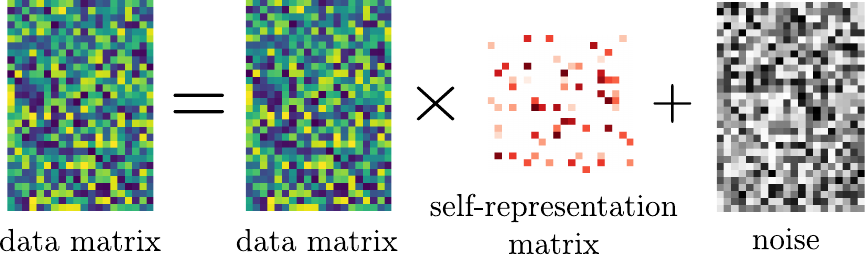}
                    \caption{An illustration of self-representation model.}
                    \label{Fig:Self-representation}
                \end{figure}

          Subspace clustering has become a major approach in analyzing high-dimensional data because it efficiently identifies meaningful low-dimensional structures within high-dimensional data. Classical model-based subspace clustering methods such as SSC~\cite{SSC} and LRR~\cite{liu2012robust} optimize self-representation-based models to unveil the affinities between different data points as shown in Fig.~\ref{Fig:Self-representation}. 
                 \begin{figure}[!t]
                    \centering
                    \includegraphics[width=0.7\linewidth]{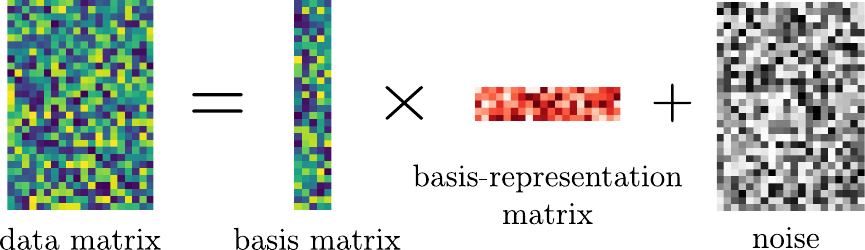}
                    \caption{An illustration of the basis-representation model.}
                    \label{Fig:Basis-representation}
                \end{figure}
          However, they rely solely on spectral representations to identify relationships between HSI pixels, making the models sensitive to noise. 
          Therefore, many extensions attempt to incorporate local and non-local information to improve robustness against noise. 
        
          Zhang et al.~\cite{7425209} proposed a spatial-spectral SSC that incorporates spatial information by applying a 2-D mean filter to the representation matrix. 
          Zhai et al.~\cite{zhai2018total} do so by introducing a total variation (TV) regularization to promote piecewise smoothness in the representation matrix. 
          Huang et al.~\cite{8643437} imposed a joint constraint on local neighborhoods obtained via super-pixel segmentation to reduce feature variability within clusters. Other methods, including~\cite{9081919,liu2020locally}, incorporate non-local structure by imposing Laplacian constraints. These extensions improved the performance of model-based subspace clustering on HSIs. However, they are limited by shallow representations that might not be discriminative enough for clustering, especially when dealing with highly complex remote-sensing images. In~\cite{bacca2017kernel, zhang2016hyperspectral}, the kernel transform is employed to nonlinearly map data into a feature space and then perform clustering in this feature space. However, the selection of kernels is largely empirical, with no clear evidence that they are well-suited for subspace clustering. 
    
        \subsection{Purely data-driven deep clustering methods}
         Data-driven deep clustering approaches focus on learning from the inherent structure and distribution of data and can be broadly categorized into two types. The first type involves feature learning followed by traditional clustering, where neural networks are used for feature extraction, after which conventional clustering algorithms are applied. For example, the deep embedding network for clustering (DEN)~\cite{DEN} employs an autoencoder with local similarity and sparsity constraints, followed by k-means clustering on the extracted features. Similarly, deep subspace clustering with sparsity prior (PARTY)~\cite{PARTY} trains an autoencoder with structure prior regularization and then applies subspace clustering. The second type is the joint optimization of feature learning and clustering, where traditional clustering algorithms are integrated into the neural network's loss function, allowing for simultaneous clustering and feature learning during network training. For instance, Xie et al.~\cite{xie2016unsupervised} propose assigning soft scores to data points based on features extracted via K-means, followed by refining these probabilities to emphasize higher values. Building on this, Nalepa et al.~\cite{8948005} employed a 3D convolutional network to better capture spectral structure, further improving performance. Another method, deep semantic clustering by partition confidence maximization (PICA)~\cite{huang2020deep}, maximizes the confidence level of each data point being assigned to a cluster, thereby enhancing clustering results. While purely data-driven deep clustering methods are flexible and can adapt to data with complex structures or noise, they often require large datasets, lack interpretability, and are prone to overfitting.

    \subsection{Model-aware deep subspace clustering}
        Model-aware deep learning, which integrates mathematical modeling with deep neural networks to harness the advantages of both domains, has been widely adopted in various image inverse problems, including image denoising~\cite{zeng2023degradation}, image reconstruction~\cite{chen2023soul}, and compressed sensing~\cite{kouni2022admm}.
        In remote sensing, this paradigm has been extensively explored through deep unfolding and plug-and-play (PnP) frameworks. For example, deep unfolding has been applied to tasks such as satellite image super-resolution~\cite{wang2022deep} and pan-sharpening~\cite{tao2023dudb}, where iterative optimization algorithms are unrolled into deep networks, offering both interpretability and improved performance. In contrast, PnP strategies embed learned priors into traditional optimization pipelines, as demonstrated in hyperspectral unmixing~\cite{zhao2021plug}.
        Similarly, model-aware deep learning techniques have shown success in subspace clustering. A pioneering work is DSCNet~\cite{DSCNet}, which employs a deep autoencoder to map data nonlinearly into a latent space and then applies a fully connected layer on the latent representation to mimic the self-representation model.
        
        To improve robustness to noise, many extensions have been proposed that incorporate non-local structure during model optimization. For example, Zeng et al.~\cite{zeng2019spectral} employ a Laplacian regularizer on the self-representation matrix to directly impose non-local structure preservation. In~\cite{cai2021graph}, the Laplacian regularizer is applied to the self-representation matrix within a residual network. Cai et al.~\cite{cai2021hypergraph} propose imposing a hypergraph regularizer on the latent representation to indirectly maintain the non-local structure in the self-representation matrix. In summary, although these methods perform well, they have several limitations. They require significant computational resources due to the large matrices involved, their two-stage design prevents integrated structure preservation, and they do not fully capture the spatial dependencies present in the data.

        Different from the DSCNet framework, the emerging basis-representation-based subspace clustering method~\cite{EDESC} attempts to learn the basis of subspaces to obtain accurate assignments of data points. 
        This method builds on the property that all vectors within a subspace can be expressed as linear combinations of that subspace’s basis vectors, as illustrated in Fig.~\ref{Fig:Basis-representation}.       
        It achieves comparable performance to self-representation-based methods in image and text-level tasks with linear complexity. 
        However, this method shows limited performance in HSI clustering, as image- and text-level clustering tasks treat each data point as an entire image or document, without considering spatial relationships, rendering spatial context irrelevant. In contrast, HSI clustering involves data points corresponding to individual pixels or image patches, where spatial continuity and non-local spectral structure are critical for accurate clustering. Our contribution is motivated by extending the basis-representation model to hyperspectral data through the integration of structural constraints that explicitly preserve both spatial continuity and non-local structure.

\section{Proposed method}
\begin{figure}[h]
        \centering
        \includegraphics[width=\textwidth]{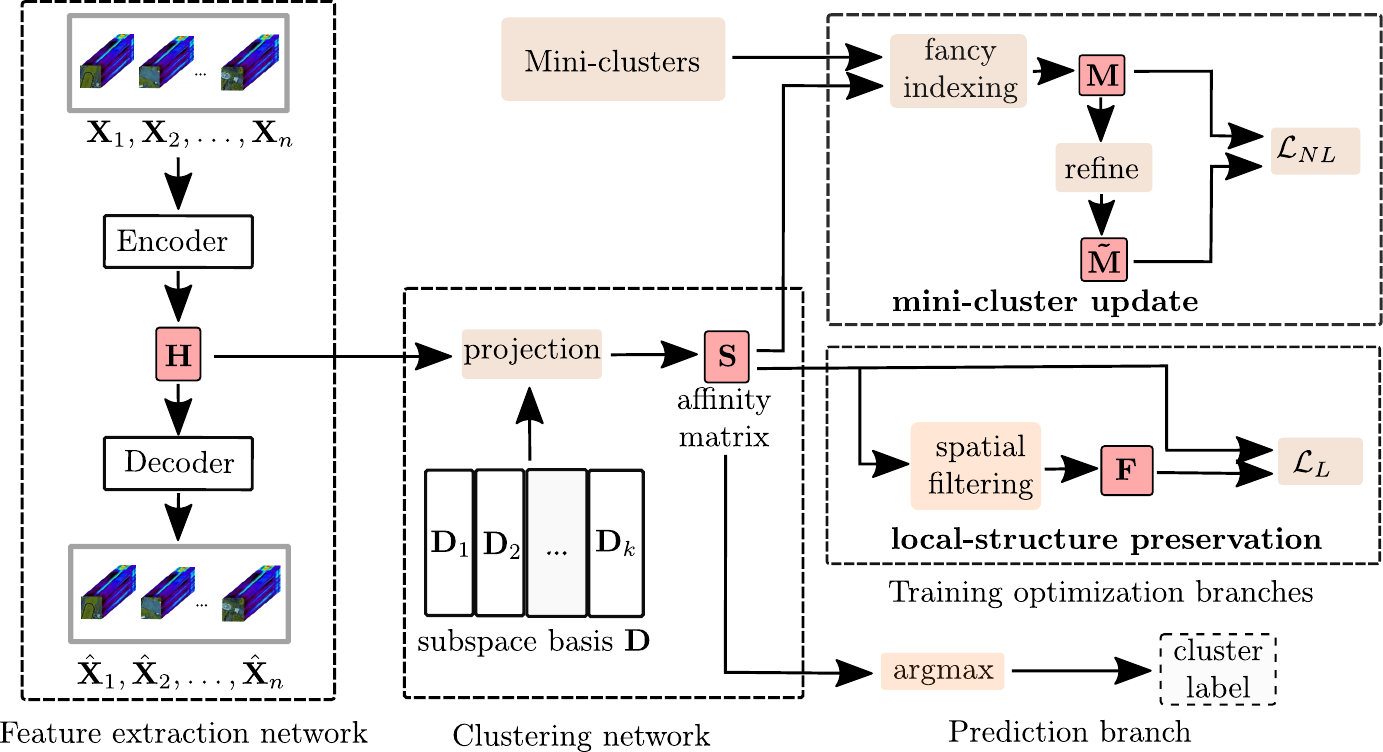}
        \caption{Structure of the proposed method. The autoencoder maps non-linearly the input data to a latent representation $\mathbf{H}$. This representation is then projected onto various subspace bases to form the subspace affinity matrix $\mathbf{S}$. 
        To enhance the quality of the subspace basis, the following optimization modules are applied: 
        (1) the mini-cluster updating module which generates mini-cluster assignment $\mathbf{M}$ and updates it by minimizing the KL loss to a refined version $\tilde{\mathbf{M}}$; 
        (2) the local-structure-preserving module which encourages the subspace affinity matrix $\mathbf{S}$ to be similar to its smooth version $\mathbf{F}$.}
        \label{Fig:full_structure}
    \end{figure}

        In this section we present our main contribution, that is a scalable and context-aware deep subspace clustering method.
        To address the high computational complexity of traditional self-representation-based clustering approaches, we propose a novel strategy that avoids learning a self-representation matrix. 
        Instead, our method directly learns a compact subspace basis, effectively integrating both local and non-local structural information inherent to HSI data.
        In this section, we first define formally the problem tackled, then we detail the structure constraints that are central to our approach, and finally we introduce our end-to-end training strategy. 
        The overall framework of our proposed method is depicted in Fig.~\ref{Fig:full_structure}.
        
        
    \subsection{Problem formulation}

        Let $\mathcal{X} = \{\mathbf{X}_1, \mathbf{X}_2, \dots, \mathbf{X}_n\}$ denote a set of $n$ HSI patches used for clustering. Each patch $\mathbf{X}_i \in \mathbb{R}^{a \times a \times b}$ is centered on a target pixel and represents a three-dimensional (3D) region of the hyperspectral image, extending over a small $a \times a$ spatial neighborhood and $b$ spectral bands. The goal of HSI clustering is to group these patches into $k$ distinct classes in an unsupervised manner. In this study, $k$ is assumed to be known in advance, which enables consistent comparison with the ground-truth labels. Note that in the subsequent discussion, the term “pixel” refers to the target center pixel of each patch, and both share the same class assignment. This patch-overlap strategy is widely adopted in hyperspectral clustering~\cite{cai2021hypergraph, cai2022superpixel}, as it enables spatial context modeling for each target pixel. Despite the induced redundancy, it provides consistent context cues that improve learning robustness. Traditional methods often rely on a two-stage framework involving an $n \times n$ self-representation matrix, which results in $\mathcal{O}(n^2)$ computational complexity. Moreover, the self-representation matrix calculation and spectral clustering are performed separately, with structure preservation applied only during the first stage, thereby limiting its impact on the overall clustering process.
        

        In this work, we propose a scalable context-aware deep subspace clustering method for HSIs. Specifically, we aim to learn the bases of different subspaces to cluster the HSI patches. To achieve optimal clustering performance on HSI data, the learned bases must satisfy the following criteria: 
        (1) each subspace should have a distinct basis; 
        (2) the bases should exhibit strong discriminative power to distinguish data points from different subspaces; 
        (3) the bases should grasp the intrinsic geometry (that is spectral properties) of the data for the non-local structure to be preserved during clustering; 
        and (4) the bases should leverage the local structure (that is spatial continuity) to enhance robustness during feature-basis alignment.
        Similarly to model-aware deep subspace clustering methods, basis learning is formulated within the latent space of an autoencoder.
        
        Based on these requirements, we formulate the following optimization problem:
        \begin{equation} 
        \min_{\hat{X}, D, H} \frac{1}{2n} \sum_{i=1}^n ||\mathbf{X}_i - \hat{\mathbf{X}}_i||_F^2 + \beta \phi(\mathbf{D}) + \beta_1 \eta(\mathbf{H}\mathbf{D}) + \beta_2 \Psi(\mathbf{H}\mathbf{D}),
        \label{problem} 
        \end{equation}
        where $\hat{\mathbf{X}}_i$ is the reconstruction version of $\mathbf{X}_i$, $\mathbf{D}={\mathbf{D}_1,\mathbf{D}_2,\dots,\mathbf{D}_k}$ is the set of bases of $k$ subspaces, each basis $\mathbf{D}_i \in \mathbb{R}^{d \times r}$ composed of $r$ basis vectors, and $\mathbf{H} \in \mathbb{R}^{n \times d}$ is the latent representation of the input data. 
        The basis dissimilarity term $\phi(\mathbf{D})$ ensures that the learned subspace bases are distinct, enabling distinguishing data points from different subspaces.
        The non-local structure preservation term $\eta(\mathbf{H}\mathbf{D})$ encourages data points to have a prediction similar to those of their nearest neighbors in the spectral space. 
        The local structure preservation term $\Psi(\mathbf{H}\mathbf{D})$ maintains spatial dependencies, ensuring that data points share similar affinities to the subspace basis as their spatial neighbors.
        %
        %
        These constraints will be discussed in detail in \Cref{basis-detail,Non-local-detail,local-detail}.
        The positive constants $\beta$, $\beta_1$, and $\beta_2$ balance the contributions of the different terms of the objective function, and their values are discussed in the experimental section (see Section~\ref{experiment}). 
        
        Based on the learned subspace bases, 
        we compute the affinity matrix \(\mathbf{S} \in \mathbb{R}^{n \times k}\) in which each entry $s_{i,j}$ corresponds to the soft assignment of the $i$th image pixel to the $j$th cluster.
        We denote by $\mathbf{s}_{i}$ the soft assignment vector of the $i$th pixel, that is the $i$th row of $\mathbf{S}$, $\mathbf{s}_{i} = [s_{i,1}, \ldots, s_{i,k}]$.
        The soft assignment for each pixel is obtained by measuring the basis alignment between the embedded representation \(\mathbf{H}\) and the subspace basis \(\mathbf{D}\) as follows:
            \begin{equation}
                s_{i,j} = \frac{\left\| \mathbf{h}_i\mathbf{D}_{j} \right\| + \theta r}{\sum_j \left( \left\| \mathbf{h}_i \mathbf{D}_{j} \right\| + \theta r \right)} ,
                \label{soft_assignment}
            \end{equation}
        where $\theta$ is a smoothing constant, and $\mathbf{D}_j$ is the basis of $j$th subspace.
        At the end of the process, each pixel is assigned to the cluster with the highest soft assignment, that is $\argmax_j s_{i,j}$ for the $i$th pixel. 
    
         In contrast to self-representation-based methods, our model does not require maintaining a self-representation matrix of size $n \times n$. Instead, it uses a basis-representation matrix of size $rk \times n$, reducing the computational complexity from $\mathcal{O}(n^2)$ to $\mathcal{O}(n)$. Moreover, our model follows a one-stage approach, where both local and non-local structure constraints jointly optimize the entire clustering process end-to-end, thereby offering stronger guidance for model optimization.
        
        \subsection{Basis dissimilarity constraint}
        \label{basis-detail}
        In subspace clustering, the bases of each subspace must be distinct, and orthogonality is often employed to reinforce this distinction, which has been shown to be beneficial for clustering performance in prior work~\cite{EDESC}. 
        Additionally, the bases are kept on the same scale to ensure more consistent and effective evaluation. This reduces overlap between subspaces, enhances the discriminative power of the bases, and ultimately leads to more accurate and robust clustering results. To ensure these properties are maintained during basis learning, we adopt a basis dissimilarity constraint $\phi(\mathbf{D})$, similar to that in \cite{EDESC}, as described below:
        \begin{equation} 
        \label{constrain} 
        \phi(\mathbf{D}) = |\mathbf{D}^\top \mathbf{D} \odot \mathbf{O}|_F^2 + |\mathbf{D}^\top \mathbf{D} \odot \mathbf{I} - \mathbf{I}|_F^2, 
        \end{equation}
        where $\odot$ represents the Hadamard product, $\mathbf{O} \in \mathbb{R}^{kr \times kr}$ is a matrix with all $r$-size diagonal block elements as 0 and all others as 1, and $\mathbf{I}$ is the identity matrix of appropriate dimensions.
        This constraint effectively enforces each subspace to have orthonormal bases (unit-norm and mutually orthogonal), and pushes bases from different subspaces apart. In other words, $\phi(\mathbf{D})$ helps $\mathbf{D}$ behave like a block-wise orthogonal basis set, enhancing subspace separation and stability.
        

    \subsection{Non-local structure preservation}
    \label{Non-local-detail}
    Image pixels with similar spectral responses are likely to belong to the same land cover class, regardless of their spatial location. Making use of these non-local spectral similarities helps to preserve the correct non-local structure in a clustering map. Existing works often apply a Laplacian matrix to maintain this structure in the self-representation matrix, which is computationally expensive and limits scalability for large datasets. 
    
    To address this issue, we propose a mini-cluster updating scheme that ensures the subspace bases align with this non-local structure of the data while efficiently preserving it in the final clustering map. Specifically, the original data points are first grouped into mini-clusters using the algorithm FINCH (see \Cref{subsec:finch} for details). 
    Let $\mathcal{C} = \{\mathcal{C}_1, \mathcal{C}_2, \mathcal{C}_3, \dots, \mathcal{C}_l\}$ represent the set of $l$ mini-clusters generated by FINCH, where each mini-cluster comprises a group of neighboring data points. 
    Formally, each mini-cluster $\mathcal{C}_p$ contains the indices of the data points belonging to this mini-cluster.
    We expect data points in each mini-cluster $\mathcal{C}_p$  to align more strongly with a common basis $D_j$ than with any other $D_e/D_j$,
    as shown below,
    \begin{equation}
        \left\| \mathbf{h}_{q}^\top \mathbf{D}_{j} \right\| \gg \underset{e \neq j}{\text{max}}\left\| \mathbf{h}_{q}^\top \mathbf{D}_e \right\| \quad \text{for all } q \in \mathcal{C}_p, 
        \label{alignment}
    \end{equation}
    where $\mathbf{h}_{q} \in \mathbb{R}^{d}$ represents the latent representation of the $q$th data point.
    According to the definition of soft assignment in~\eqref{soft_assignment}, the subspace basis affinity in~\eqref{alignment} can be mapped to soft assignments, resulting in similar assignments for all data points within the same mini-cluster, that is for a given mini-cluster $\mathcal{C}_p$,
     \begin{equation}
        \mathbf{s}_{q} \approx \mathbf{s}_{o} \text{ for all } q \text{ and } o \in \mathcal{C}_p .
    \end{equation}
    To preserve the underlying structure, we optimize the soft-assignment at the mini-cluster level to encourage data points within the same mini-cluster to share the same assignment. During this processing, the soft assignments of data points within each mini-cluster are extracted using their mini-cluster index through fancy indexing—a vectorized technique that enables efficient, loop-free index-based value extraction on GPUs. The soft assignments of mini-clusters are represented as $\mathbf{M} \in \mathbb{R}^{l \times k}$, $\mathbf{M} =\{\mathbf{m}_1, \mathbf{m}_2,\dots,\mathbf{m}_{l}\}$. The soft assignment for the $p$th mini-cluster $\mathbf{m}_p  \in \mathbb{R}^{k}$ is obtained by 
    averaging the assignments of the data points it contains, as follows,
    \begin{equation}
        \mathbf{m}_p = \frac{1}{|\mathcal{C}_p|}\sum_{q \in \mathcal{C}_p}\mathbf{s}_{q}.
    \end{equation}
    To further enhance the distribution of these assignments, we adopt a refined soft assignment $\tilde{\mathbf{M}} \in \mathbb{R}^{l \times k}$ whose entries are defined as
    \begin{equation}
        \tilde{m}_{p,j} = \frac{m_{p,j}^{2}/\sum_{p}m_{p,j}}{\sum_{j}(m_{p,j}^{2}/\sum_{p}m_{p,j})},
    \end{equation}
    where $m_{p,j}$ represents the soft assignment of the $p$th mini-cluster to class $j$, and $\tilde{m}_{p,j}$ is its refined soft-assignment. 
    This refinement process improves cluster purity by emphasizing high-confidence predictions and mitigating distortions caused by large clusters~\cite{xie2016unsupervised}. By aligning the initial mini-cluster predictions with their refined versions, the quality of soft assignments is enhanced, which in turn strengthens the discriminative power of the subspace bases. Based on the above analysis, we define the non-local structure preservation constraint $\eta(\mathbf{H}\mathbf{D})$ as follow:
    \begin{equation}
        \eta(\mathbf{H}\mathbf{D}) = \kl(\tilde{\mathbf{M}}||\mathbf{M}) = \sum_{p}\sum_{j}\tilde{m}_{p,j}\log{\frac{\tilde{m}_{p,j}}{m_{p,j}}},
    \end{equation} 
    where the $\tilde{\mathbf{M}}$ represents the refined soft assignment affinity matrix of mini-cluster. During training, the refinement calculation increasingly emphasizes class distinctions, causing the refined mini-cluster soft assignment $\tilde{\mathbf{m}}_i$ to converge towards a state where one class dominates, with its value approaching 1, while the values for other classes approach 0. As $\eta(\mathbf{H}\mathbf{D})$ decreases over time, the mini-cluster is assigned more confidently to a single class.

    In parallel, the soft-assignment of data points within the mini-cluster follows the same optimization trajectory as the mini-cluster itself as follows:
    \begin{equation}
        \frac{\partial \eta(\mathbf{H}\mathbf{D})}{\partial \mathbf{s}_{q}} = \frac{\partial \eta(\mathbf{H}\mathbf{D})}{\partial \mathbf{m}_{p}} \frac{\partial \mathbf{m}_{p}}{\partial \mathbf{s}_{q}} = \frac{1}{|\mathcal{C}_{p}|} \frac{\partial \eta(\mathbf{H}\mathbf{D})}{\partial \mathbf{m}_{p}},\quad \text{for all } q \in \{\mathcal{C}_p\}.
        \label{grandient}
    \end{equation}   
     This means that as the mini-cluster moves toward being classified as a particular class, all the data points within it also shift toward the same class assignment. As training progresses, this optimization direction causes all the points in the mini-cluster to converge to the same class assignment.
    The mini-cluster updating scheme enhances the model's performance in two ways.
    First, it optimizes the mini-cluster soft assignments, which are more representative and robust than instance assignment, because of the assignment averaging.
    Second, it promotes consistency among data points within the same mini-cluster, ensuring they share the same prediction and preserving the non-local structure during optimization. 
    Unlike previous methods that relied on a Laplacian matrix with complexity $\mathcal{O}(n^2)$ to preserve the non-local spatial structure, our approach is significantly more efficient. By leveraging a mini-cluster index vector of size $n$ and using fancy indexing with complexity $\mathcal{O}(n)$ aligns with GPU implementation efficiency, we achieve efficient non-local structure preservation. Moreover, our method integrates non-local structure preservation throughout the entire clustering process, rather than limiting them to a single stage. This provides stronger guidance and improves the overall clustering performance.

    \subsection{Local structure preservation}
    \label{local-detail}
\begin{figure}[ht]
            \centering
            \includegraphics[width=0.9\linewidth]{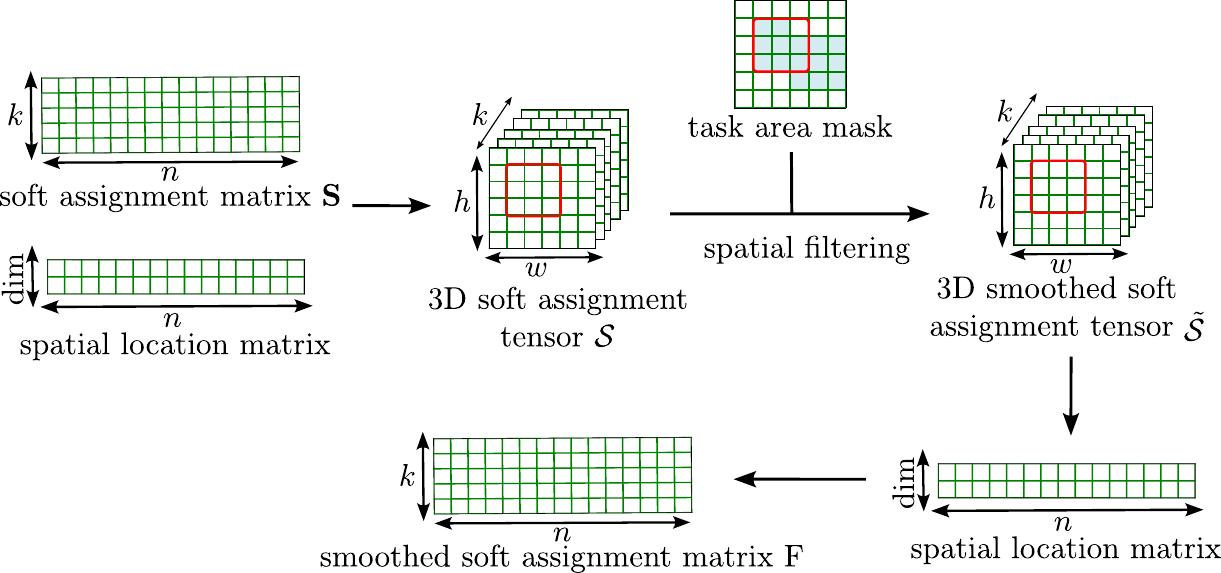}
            \caption{Calculation process of the smoothed soft assignment matrix $\mathbf{F}$. First, the soft assignments of image pixels are arranged into a 3D tensor based on their spatial locations. Next, a local window is applied to perform mean filtering for each point. Specifically, the filtering is calculated within the task area defined by a MASK $\mathbf{T}$, where labeled data is available, to facilitate effective evaluation of the results.}
            \label{smooth-prediction}
        \end{figure}

         In real-world scenarios, neighboring areas often belong to the same land-cover class, a property known as spatial continuity. This relationship is a common phenomenon in many types of data, including HSIs, where adjacent pixels are likely to belong to the same class. To leverage this property, our model integrates spatial neighborhood information into the basis learning process. This not only improves noise robustness when measuring the alignment between feature vectors and basis vectors but also ensures that the clustering results maintain the intrinsic spatial continuity of HSIs. Specifically, as with the non-local structure preservation constraint, we directly apply spatial filtering to the soft-assignments of pixels, as illustrated in Fig.~\ref{smooth-prediction}.

        As defined in the figure, $\mathcal{S} \in \mathbb{R}^{w \times h \times k}$ represents a 3D tensor derived from the 2D soft assignment matrix, where $w$ and $h$ are the width and height of the HSI. Each $\mathcal{S}_{x,y}\in \mathbb{R}^{k}$ denotes the soft assignment vector for the pixel at spatial location $(x, y)$. By incorporating spatial information, this tensor facilitates a spatially-aware prediction of cluster membership. A spatial filtering operation is subsequently applied to each layer of $\mathcal{S}$, leveraging relationships among neighboring pixels to produce a smoothed 3D tensor $\tilde{\mathcal{S}}$, expressed as:

        \begin{equation}
        \tilde{\mathcal{S}}_{x,y} = \frac{\sum_{(u, v) \in \mathbf{W}_{x,y}} t_{u,v} \cdot \mathcal{S}_{u,v}}{\sum_{(u, v) \in \mathbf{W}_{x,y}} t_{u,v}},
        \end{equation}
        
        Where \(\tilde{\mathcal{S}}_{x,y}\) is the smoothed soft assignment vector at spatial location \((x, y)\), with \(x\) and \(y\) denoting the row and column indices, respectively, and \(\mathbf{W}(x, y)\) is a fixed smoothing window centered at \((x, y)\). The mask \(\mathbf{T}\in \mathbb{R}^{w \times h}\)  assigns a value of 1 to pixels within the cluster region and 0 otherwise, where \(t_{x,y}\) denotes the value at the corresponding location in \(\mathbf{T}\). We then extract the final smoothed soft-assignment matrix $F\in \mathbb{R}^{n×k}$ by flattening S~ according to pixel ordering.

        To incorporate this refined local structure into the original predictions, we minimize the Kullback-Leibler~(KL) divergence between the original and the smoothed predictions. The local structure preservation function $\Psi(\mathbf{H}\mathbf{D})$ is defined as follows:
        \begin{equation}
            \Psi(\mathbf{H}\mathbf{D}) = \kl(\mathbf{F}||\mathbf{S}) = \sum_{i}\sum_{j}{f}_{i,j}\log{\frac{{f}_{i,j}}{s_{i,j}}},
        \end{equation}
        where $\mathbf{F}\in \mathbb{R}^{n \times k}$ is the smoothed assignment matrix, $f_{i,j}$ represents the element at position $(i, j)$ and $\mathbf{S}\in \mathbb{R}^{n \times k}$ is the original soft-assignment matrix.  

        Existing methods typically apply spatial constraints such as total variation (TV) on a self-representation 3D tensor $\mathcal{M} \in \mathbb{R}^{w \times h \times n^2}$, where $w$ and $h$ are the width and height of the HSI, resulting in a computational complexity of $\mathcal{O}(n^2)$. 
        In contrast, our method directly applies spatial filtering to the clustering soft-assignment $\mathcal{S} \in \mathbb{R}^{w \times h \times k}$. Denoting the size of the smoothing window as $|\mathbf{W}|$, the computational complexity is $\mathcal{O}(|\mathbf{W}| \times n \times k)$, which simplifies to $\mathcal{O}(n)$, making it much more efficient. More importantly, our spatial constraint optimizes the entire clustering process, providing stronger guidance for network optimization.

        \subsection{Objective function and training strategy}
        The objective function consists of multiple loss terms that jointly optimize the reconstruction of the autoencoder, basis dissimilarity, and both local and non-local structure preservation. The reconstruction loss is defined as:
        \begin{equation}
           \mathcal{L}_{R}  =  \frac{1}{2n} \sum_{i=1}^n ||\mathbf{X}_i - \hat{\mathbf{X}}_i||_F^2.
        \end{equation}        
        According to the definition of $\phi(\mathbf{D})$, the basis dissimilarity loss is formulated as:
        \begin{equation}
           \mathcal{L}_{D} = \phi(\mathbf{D}) 
        \end{equation} 
        The non-local preservation loss, based on the definition of $\eta(\mathbf{H}^\top \mathbf{D})$, is formulated as:
        \begin{equation}
            \mathcal{L}_{NL} = \eta(\mathbf{H}^\top \mathbf{D})
        \end{equation}
        Similarly, the local structure preservation loss is defined as:
        \begin{equation}
            \mathcal{L}_{L} = \Psi(\mathbf{H}^\top \mathbf{D})
        \end{equation}
        The total objective function is the weighted sum of these loss terms:
        \begin{equation}
            \mathcal{L}_{\text{total}} = \mathcal{L}_{R} + \beta \cdot \mathcal{L}_{D} + \beta_1 \cdot \mathcal{L}_{NL} + \beta_2 \cdot \mathcal{L}_{L}.
        \end{equation}
        
        \begin{algorithm}[!h]
        \caption{Training Process for Hyperspectral Image Clustering.}
        \begin{algorithmic}[1]
        \State \textbf{Input:} Hyperspectral image patches: $\mathcal{X} = \{\mathbf{X}_1, \mathbf{X}_2, \dots, \mathbf{X}_n\}$
        \State \textbf{Output:} Cluster labels 
        
        \State \textbf{Step 1: Mini-cluster generation (FINCH)}
        \State Generate mini-clusters: $\mathcal{C} = \{\mathcal{C}_1, \mathcal{C}_2, \mathcal{C}_3, \dots, \mathcal{C}_l\}$
        
        \State \textbf{Step 2: Pre-training}
        \State Perform data preprocessing and pre-train the deep autoencoder.
        \State Minimize the reconstruction loss:
        \begin{equation*}
            \mathcal{L} = \mathcal{L}_{R}
        \end{equation*}
        
        \State \textbf{Step 3: Initial subspace basis construction}
        \State Apply K-means clustering on the deep features to generate initial clusters.
        \State Initialize the subspace basis for the initial clusters using Singular Value Decomposition (SVD).

        \State \textbf{Step 4: Subspace basis refinement}
        \State Train the network with all loss components together:
        \begin{equation*}
            \mathcal{L} = \mathcal{L}_{\text{total}}
        \end{equation*}

        \State Assign each data point to a cluster based on the highest value in its soft assignment vector:
        \begin{equation*}
            \text{label}_{i} = \argmax(\mathbf{S}_i)
        \end{equation*}
        \end{algorithmic}
        \label{train-strategy}
        \end{algorithm}

     As shown in algorithm~\ref{train-strategy}, the network initialization involves 4 steps. First, the mini-clusters are generated by FINCH. Next, the autoencoder is pre-trained to obtain initial latent representations of the input data. At last, the K-means algorithm is applied to generate the initial clustering results, which are subsequently processed using Singular Value Decomposition (SVD) to obtain the initial subspace basis for each cluster. Following the work~\cite{EDESC}, we select 5 main basis vectors, i.e., the top 5 singular vectors with the largest singular values, to capture the subspace structure within each cluster. In the final step, the autoencoder and the subspace basis are jointly optimized in the final training phase, where their parameters are updated under the supervision of all constraints.

     Our model jointly optimizes these components, enabling local homogeneity to propagate to non-local data points through the non-local constraint, and vice versa. In contrast, self-representation-based subspace clustering methods have a complexity of $\mathcal{O}(n^2)$ for computing the self-representation matrix, maintaining the non-local structure with Laplacian regularization, and preserving spatial dependency on the self-representation matrix. Our method achieves a complexity of $\mathcal{O}(n)$ for all key operations, including calculating the soft assignment, maintaining the non-local structure, and preserving the local structure, making it scalable for large datasets. Furthermore, the proposed structure constraints supervise the entire clustering process, providing stronger guidance for optimization. 
     
\section{Experiments and results}
\label{experiment}

In this section, we evaluate the proposed method on 3 real-world hyperspectral images. We compare it against several popular clustering algorithms. Then we perform an ablation study to understand the impact of the local and non-local spatial structure preservation constraints.

\subsection{Datasets}
    We conducted experiments on three real-world hyperspectral image datasets. The details of these datasets are as follows:
            \begin{enumerate}
                \item \textit{Trento}:
                    This dataset was acquired using the CASI sensor and contains 63 spectral bands. It is divided into 6 classes. The image size is 600$\times$166 pixels, with 30,214 labeled samples.
                
                \item \textit{Houston}:
                    The Houston dataset was collected using the ITRES-CASI sensor and encompasses 144 spectral bands. It is categorized into 7 classes. The image size is 130$\times$130 pixels, containing 6,104 labeled samples.
                
                \item \textit{PaviaU}:
                    The PaviaU dataset was obtained using the ROSIS-3 sensor and includes 103 spectral bands. It is classified into 9 classes. The image size is 610$\times$340 pixels, with 42,776 labeled samples.
            \end{enumerate}
    These datasets feature a variety of sensor types, numbers of spectral bands, class categories, and image dimensions, providing a diverse experimental platform to validate the effectiveness of our proposed methods.

\subsection{Experimental setting}
    During training, for the Houston and Trento datasets, we use a fixed learning rate of $5 \times 10^{-3}$, while for PaviaU, a smaller rate of $1 \times 10^{-4}$ is adopted to ensure stable convergence. The model is trained for 400 epochs without early stopping. To balance memory consumption and enrich the feature representation of each HSI pixel, we use a patch size of \(7 \times 7\) for all compared methods. When generating mini clusters, we employ a larger patch size of \(17 \times 17\) to incorporate more detailed neighborhood information. Parameter $\beta$ is fixed at $10^{-3}$.  All implementation details, including hyperparameter settings and training schedule, are available in our released code.  
    
    We compare our method with several clustering approaches, including centroid-based methods such as K-means ~\cite{Kmeans} and Fuzzy C-means ~\cite{FCM}; the graph-based method Spectral Clustering ~\cite{spectral_clustering}; data-driven deep clustering methods like IDEC ~\cite{guo2017improved}, SpectralNet~(SN)~\cite{SpectralNet}, and DEKM ~\cite{DEKM}; and the nearest neighbors-based method FINCH ~\cite{finch}. Additionally, we also compare it with our preliminary work ~\cite{li2023model}, which preserves non-local structures through contrastive learning. 
    
    To assess the performance of our model, we employ three widely used metrics: Overall Accuracy~(OA), Normalized Mutual Information~(NMI), and Cohen’s Kappa~($\mathcal{K}$). The OA quantifies the proportion of correctly classified samples relative to the total number of samples, computed by:
    \begin{equation}
        \text{OA} = \frac{1}{n} \sum_{i=1}^{n} \mathbb{I}(g_i = \hat{g}_i)
    \end{equation}
    where $n$ is the total number of samples, \(y_i\) is the true label of the $i$th sample, $\hat{y}_i$ is the predicted label of the $i$th sample, and \(\mathbb{I}(\cdot)\) is the indicator function that equals 1 if the condition inside is true and 0 otherwise. The NMI measures the mutual information between the clustering results and the true labels, normalized by the average of their entropies, defined by:
    \begin{equation}
        \text{NMI} = \frac{2 \times \text{I}(\mathbf{G}; \hat{\mathbf{G}})}{\text{H}(\mathbf{G}) + \text{H}(\hat{\mathbf{G}})}
    \end{equation}
    where $\text{I}(\mathbf{G}; \hat{\mathbf{G}})$ is the mutual information between the true labels \(\mathbf{G}\) and the clustering labels \(\hat{\mathbf{G}}\), and \(\text{H}(\mathbf{G})\) and \(\text{H}(\hat{\mathbf{G}})\) are the entropies of \(\mathbf{G}\) and \(\hat{\mathbf{G}}\), respectively. $\mathcal{K}$ assesses the agreement between the clustering results and the true labels while accounting for the possibility of agreement occurring by chance, calculated by:
    \begin{equation}
        \mathcal{K} = \frac{P_o - P_e}{1 - P_e}
    \end{equation}
    where \(P_o\) is the observed agreement among raters, and \(P_e\) is the expected agreement by chance. For the three evaluation metrics, a higher value indicates a better performance. 

    \subsection{Performance analysis}
    \subsubsection{Houston dataset:} 
\begin{figure}[ht]
    \centering
    \includegraphics[width=\linewidth]{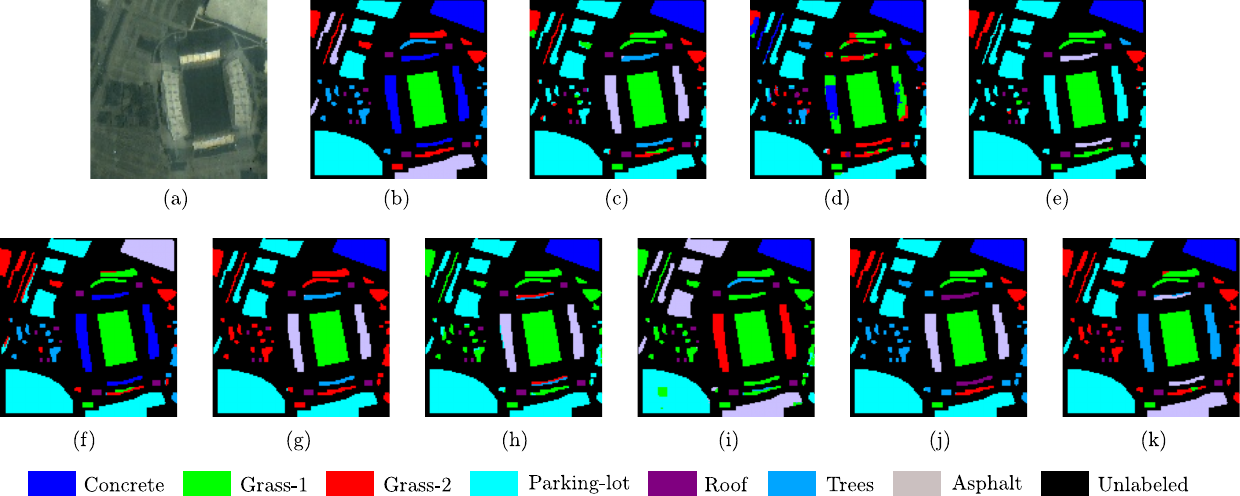}
    \caption{Houston dataset. (a) False-color image. (b) Ground truth and the clustering results obtained by (c) k-means, (d) FCM, (e) SC, (f) FINCH,
(g) IDEC, (h) DEKM, (i) SpectralNet, (j) MADL and (k) SCDSC.}
    \label{Visual_Houston}
\end{figure}

The clustering results of different methods on the Houston dataset are presented in Table~\ref{Houstonresult} and Fig.~\ref{Visual_Houston}. 
    \begin{table}[h]
    \centering
    \setlength{\tabcolsep}{4pt}
    \resizebox{\textwidth}{!}{
        \begin{tabular}{c|ccccccccc}
            \hline
            Class & Kmeans\cite{Kmeans} & FCM\cite{FCM} & SC\cite{spectral_clustering} & IDEC~\cite{guo2017improved} & FINCH~\cite{finch} & DEKM~\cite{DEKM} & SN~\cite{SpectralNet} & MADL~\cite{li2023model} & SCDSC \\
            \hline 
            1    & 46.50 & \textbf{85.04} & \underline{62.46} & 47.20 & 53.50 & 45.53 & 41.40 & 47.20 & 48.60 \\ 
            2     & \textbf{100} & \underline{75.91} & \textbf{100} & \textbf{100} & \textbf{100}  & \textbf{100} & \textbf{100}  & \textbf{100} & \textbf{100} \\ 
            3     & 58.05 & 16.15 & 35.66 & \textbf{88.94} & 70.61 & 24.30 & 32.11 & 74.70 & \underline{79.66} \\ 
            4 & \textbf{100} & 99.81 & \textbf{100} & 96.37 & \textbf{100} & \textbf{100}  & 76.44 & \underline{99.96} & 99.65 \\ 
            5        & \underline{94.77} & 15.77 & 90    & 94.54 & \textbf{100} & 85.69 & 80    & 69.38 & 60    \\ 
            6       & 0      & 27.07 & 0      & 0      & \textbf{70.94} & 0      & 1.54 & 23.05 & \underline{30.70} \\ 
            7     & 0      & 3.87 & 0      & 9.97 & 0      & 4 & \textbf{59.16} & 39.56 & \underline{47.97} \\ 
            \hline
            OA($\%$)   & 64.50 & 66.33 & 65.86 & 67.58 & 72.10 & 61.01 & 60.77 & \underline{72.33} & \textbf{74.41} \\ 
            NMI    & 0.6973 & 0.6204 & 0.6181 & \underline{0.7851} & 0.7702 & 0.7011 & 0.6439 & 0.7656 & \textbf{0.7902} \\ 
            $\mathcal{K}$       & 0.5354 & 0.5531 & 0.5441 & 0.5859 & 0.6424 & 0.4922 & 0.5057 & \underline{0.6492} & \textbf{0.6759} \\ 
            \hline
        \end{tabular}
    }
    \caption{Quantitative evaluation of different clustering methods on the dataset Houston}
    \label{Houstonresult}
\end{table}
    In Table~\ref{Houstonresult}, the best result is highlighted in \textbf{bold}, and the second best is \underline{underlined}. We set $\beta_1 = 3$, $\beta_2 = 8$, and use a filter window size of $3 \times 3$. 
    We observe that our method achieved the best performance in terms of OA, NMI, and $\mathcal{K}$. Compared to our preliminary work~\cite{li2023model}, our model achieves an additional $2.08\%$ improvement in OA, demonstrating the advantage of the mini-cluster updating scheme. Notably, two purely data-driven deep clustering methods, DEKM and SpectralNet, obtained the lowest accuracies, possibly because the data volume is relatively small, which limited the learning capability of deep neural networks. Compared to centroid-based methods such as Kmeans and FCM, and graph-based methods like Spectral Clustering, the FINCH algorithm showed much better performance, indicating its powerful ability to handle complex structures using a nearest neighbors chain approach. From the cluster map, we observe that our method best aligns with the ground truth. Specifically, our approach accurately distinguishes between parking-lot and asphalt areas. In contrast, other methods predominantly cluster these regions as parking-lot. Only SpectralNet partially recognizes these distinctions.
    
    \subsubsection{Trento dataset:} 
\begin{figure}[ht]
    \centering
    \includegraphics[width=\linewidth]{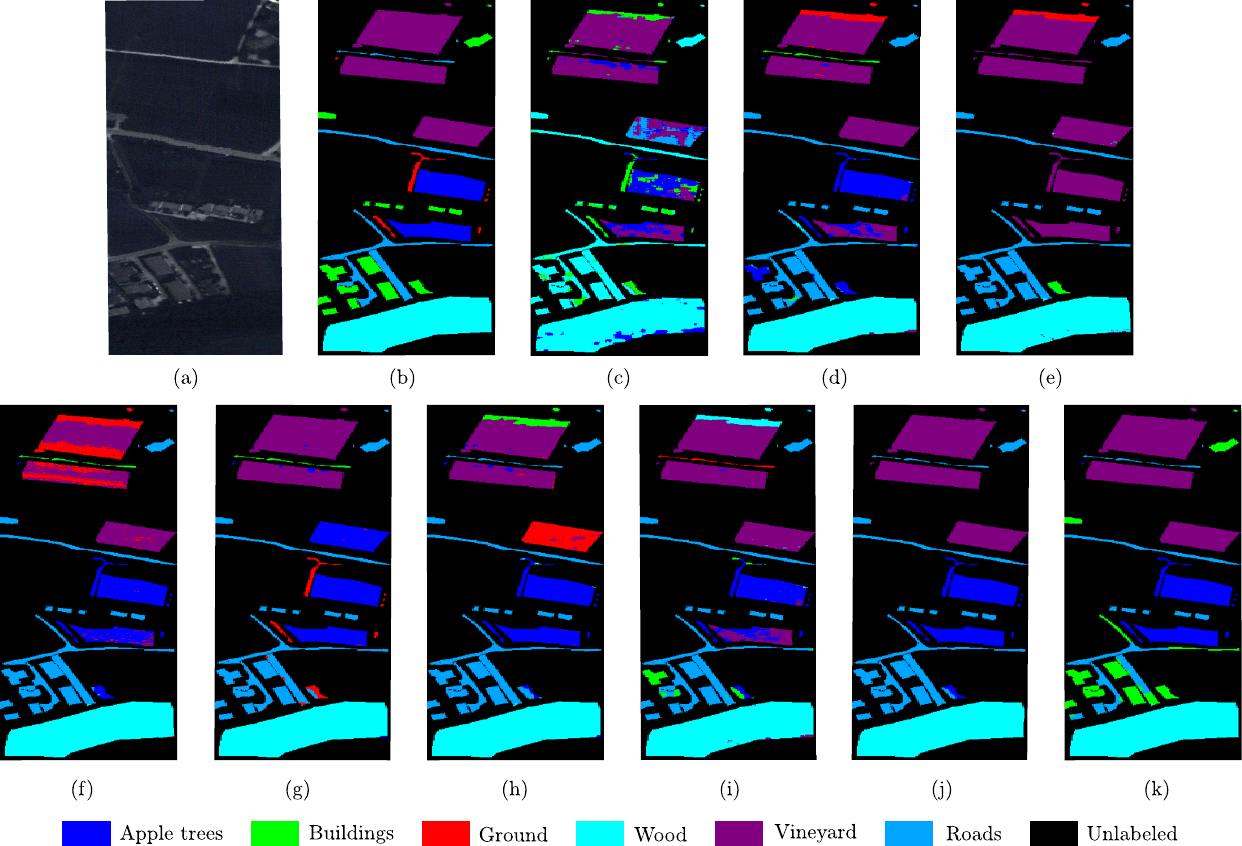}
    \caption{Trento dataset. (a) False-color image. (b) Ground truth and the clustering results obtained by (c) k-means, (d) FCM, (e) SC, (f) FINCH, (g) IDEC, (h) DEKM, (i) SpectralNet, (j) MADL and (k) SCDSC.}
    \label{Visual_Trento}
\end{figure}

The clustering results of various methods on the Trento dataset are presented in Table~\ref{Trentoresult} and Fig.~\ref{Visual_Trento}.
    \begin{table}[!h]
    \centering
    \setlength{\tabcolsep}{4pt}
    \resizebox{\textwidth}{!}{
        \begin{tabular}{c|ccccccccc}
            \hline
            Class & Kmeans\cite{Kmeans} & FCM\cite{FCM} & SC\cite{spectral_clustering} & IDEC~\cite{guo2017improved} & FINCH~\cite{finch} & DEKM~\cite{DEKM} & SN~\cite{SpectralNet} & MADL~\cite{li2023model} & SCDSC \\
            \hline      
            1  & 71.37 & 20.86 & 0      & 89.50 & 93.95 & 99.39 & 73.69 & \underline{99.98} & \textbf{100} \\
            2    & 14.33 & 6.22  & 7.82  & 35.38 & 0      & 31.85 & \underline{39.84} & 10    & \textbf{87.98} \\
            3      & 0      & \underline{3.36}  & 0      & \textbf{53.90} & 0      & 0      & 0      & 0      & 0      \\
            4        & 99.29 & 83.25 & 99.52 & 99.16 & 99.54 & \underline{99.91} & 99.41 & \textbf{100}  & \textbf{100}  \\
            5    & 92.19 & 85.75 & 95.54 & 76.92 & 91.96 & 76.44 & \underline{92.55} & \textbf{100}  & \textbf{100} \\
            6        & 84.72 & 54.77 & \textbf{92.82} & 70.47 & 67.13 & 80.04 & 69.99 & \underline{86.05} & 36.80 \\
            \hline
            OA($\%$)         & 81.83 & 64.13 & 73.76 & 80.28 & 76.23 & 81.47 & 83.20 & \underline{88.30} & \textbf{90.61} \\ 
            NMI         & 0.7717 & 0.5900 & 0.7621 & 0.8234 & 0.8200 & 0.8257 & 0.7835 & \textbf{0.9144} & \underline{0.9101} \\ 
            $\mathcal{K}$       & 0.7566 & 0.5049 & 0.6353 & 0.7430 & 0.6963 & 0.7607 & 0.7720 & \underline{0.8434} & \textbf{0.8746} \\
            \hline
        \end{tabular}
    }
    \caption{Quantitative evaluation of different clustering methods on the dataset Trento}
    \label{Trentoresult}
\end{table}

    We set $\beta_1 = 3$, $\beta_2 = 1$, and use a filter window size of $7 \times 7$. We observe that most methods achieve accuracy rates exceeding $70\%$, which may be attributed to data imbalance, as the majority of data points belong to a few dominant classes. Meanwhile, due to the class imbalance, most methods including ours failed to recognize the 3rd class. Our method achieves the best performance in terms of OA and $\mathcal{K}$, with only a slight decrease in NMI. Compared to our preliminary work in~\cite{li2023model}, our model demonstrates an additional $2.32\%$ improvement in OA while maintaining an NMI that is just $0.43\%$ lower. On this dataset, the three purely data-driven deep clustering methods outperform most conventional clustering methods, likely benefiting from the large volume of labeled data. However, K-means also performs exceptionally well, and significantly better than FCM. This superior performance of K-means can be attributed to the clear boundaries in the data, which make hard assignments more effective. Furthermore, K-means outperforms FINCH and Spectral Clustering, possibly because the dataset exhibits a structure that is close to a spherical shape. From the cluster map, we observe that our method aligns with the ground truth the most. Specifically, our method accurately recognizes the Vineyard and Wood classes, whereas other methods tend to mix these classes with others. More importantly, our method produces much smoother predictions compared to other methods. Compared to our preliminary work, the new non-local structure preservation scheme enhances the ability to distinguish buildings and roads.

\subsubsection{PaviaU dataset}
\begin{figure}[ht]
    \centering
    \includegraphics[width=\linewidth]{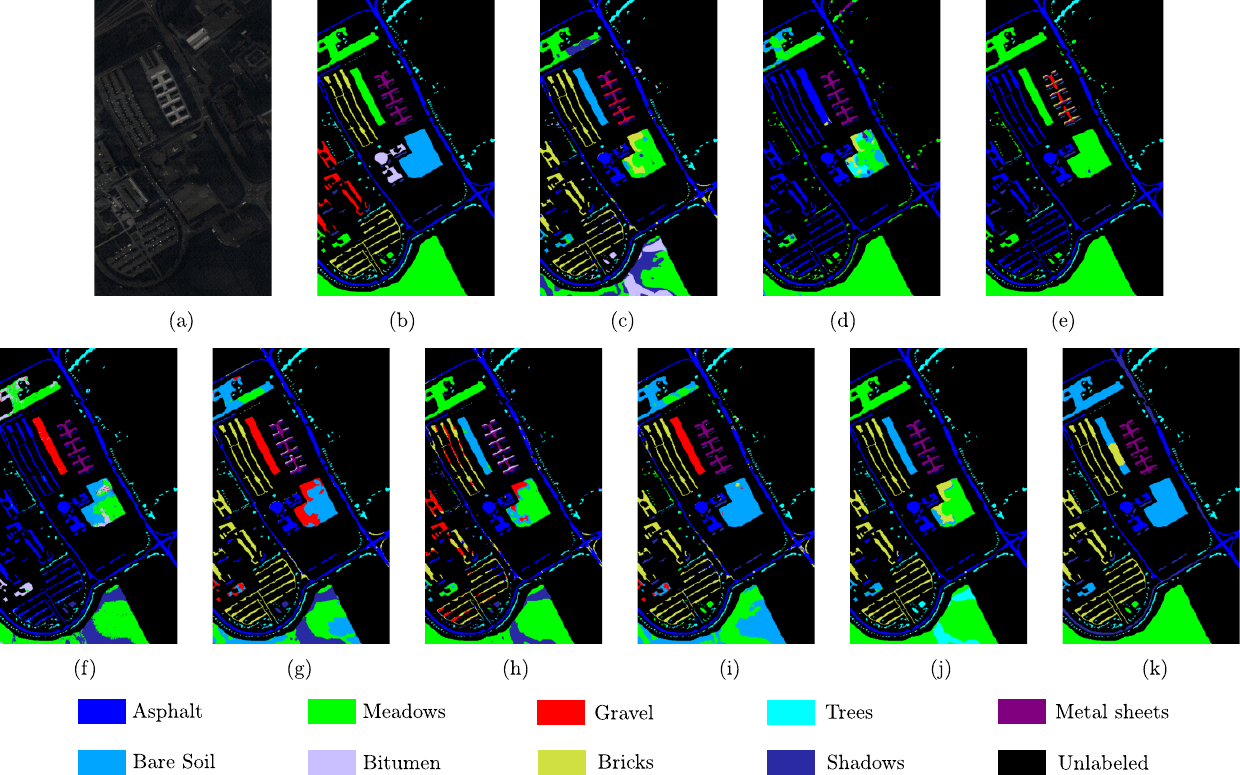}
    \caption{PaviaU dataset. (a) False-color image. (b) Ground truth and the clustering results obtained by (c) k-means, (d) FCM, (e) SC, (f) FINCH,
(g) IDEC, (h) DEKM, (i) SpectralNet, (j) MADL and (k) SCDSC.}
    \label{Pavia_visualize}
\end{figure}

    The clustering results of various methods on the PaviaU dataset are presented in Table~\ref{Paviaresult} and Fig.~\ref{Pavia_visualize}. 
    \begin{table}[!h]
    \centering
    \setlength{\tabcolsep}{4pt}
    \resizebox{\textwidth}{!}{
        \begin{tabular}{c|ccccccccc}
            \hline
            Class & Kmeans\cite{Kmeans} & FCM\cite{FCM} & SC\cite{spectral_clustering} & IDEC~\cite{guo2017improved} & FINCH~\cite{finch} & DEKM~\cite{DEKM} & SN~\cite{SpectralNet} & MADL~\cite{li2023model}& SCDSC  \\
            \hline
            1     & 92.44 & 94.30 & \textbf{100}     & 96.53 & \underline{99.98} & 77.04 & 95.17 & 99.55 & 93.64 \\
            2     & 47.75 & 73.85 & \textbf{100}     & 38.48 & 58.88 & \underline{75.33} & 63.97 & 70.83 & 57.67 \\
            3       & 0      & \underline{3.89} & 0      & 0      & 0      & \textbf{10.57} & 0      & 0      & 0      \\
            4      & 72.63 & 56.25 & 73.76 & 95.59 & 89.49 & 65.43 & 77.44 & \underline{97.24} & \textbf{98.98} \\
            5       & 66.40 & 46.35 & 39.18 & 76.68 & \textbf{100}      & 93.42 & \textbf{100}      & \underline{99.96} & \textbf{100}     \\
            6      & 3.92 & 15.90 & 0      & \underline{53.16} & 44.04 & 22.13 & 14.94 & 18.65 & \textbf{85.46} \\
            7          & 0      & 0      & 0      & 0      & 0      & \textbf{17.32} & \underline{0.28} & 0      & 0      \\
            8     & 93.69 & 18.13 & 0      & \underline{99.33} & 0      & 82.31 & 54.66 & 79.84 & \textbf{99.98} \\
            9     & 0      & 30.22 & \underline{76.24} & 14.20 & 0      & 68.00 & \textbf{94.41} & 9.10 & 42.20 \\
            \hline
            OA($\%$) & 50.97 & 56.59 & \underline{67.30} & 56.11 & 55.90 & 64.66 & 59.89 & 65.69 & \textbf{69.48} \\
            NMI         & 0.5926 & 0.4626 & \underline{0.6905} & 0.6722 & 0.6510 & 0.6480 & 0.6408 & 0.6511 & \textbf{0.7490} \\
            $\mathcal{K}$  & 0.3918 & 0.4257 & 0.5290 & 0.4811 & 0.4762 & 0.5408 & 0.4805 & \underline{0.5521} & \textbf{0.6250} \\
            \hline
        \end{tabular}
    }
    \caption{Quantitative evaluation of different clustering methods on the dataset PaviaU}
    \label{Paviaresult}
\end{table}

    We set $\beta_1 = 3$, $\beta_2 = 7$, and use a filter window size of $7 \times 7$. Our method achieves the best performance in terms of OA, NMI, and $\mathcal{K}$. 
    Compared to our preliminary work in~\cite{li2023model}, our model demonstrates an improvement of $3.79\%$ in OA.
    Due to the dataset complex structure, all methods attain an OA below $70\%$. Spectral Clustering achieves the second-best result, outperforming many deep learning methods despite the large data volume. This superior performance is likely because the data possess a complex non-local structure that graph-based methods can model more effectively than others. However, the Spectral clustering fails to recognize 4 of 9 classes. Our method also fails to recognize 2 classes, maybe because of the serious class imbalance. Generally, the data-driven deep clustering methods perform better than conventional methods in this context. From the cluster map, we observe that our method best aligns with the ground truth. Specifically, our approach accurately distinguishes meadows and trees, whereas other methods often confuse these classes. Furthermore, benefiting from local structure preservation, both our method and the preliminary work exhibit smoother predictions.

    \subsection{Ablation study}
    In the ablation study, we analyze the effect of different parts of our model. Specifically, we applied the basis-representation clustering model proposed in~\cite{EDESC} as the baseline. We then enhanced this model by introducing a mini-cluster-based update scheme to preserve the spectral structure. Finally, we integrated a local preservation module to smooth out noise further and improve robustness. The parameter setting of our model is shown in Table~\ref{Hyperparametersetting}:
    \begin{table}[h]
        \centering
            \begin{tabular}{ccccc}
                \toprule
                        dataset&learning rate&$\beta_1$&$\beta_2$&smooth window\\
                \midrule
                        Houston &   0.0001   &   3   &   8   &  $3\times3$\\
                        Trento  &   0.005    &   3   &   1   &  $7\times7$\\
                        PaviaU  &   0.005    &   3   &   7   &  $7\times7$\\
                \bottomrule
            \end{tabular}
        \caption{Hyperparameter setting.}
        \label{Hyperparametersetting}
    \end{table}
    The results are presented in Table~\ref{tab:ablation_study}:
    \begin{table}[!h]
        \centering
        \resizebox{\textwidth}{!}{%
            \begin{tabular}{cccccccccc}
                \toprule
                \multirow{2}{*}{Model} & \multicolumn{3}{c}{Houston} & \multicolumn{3}{c}{Trento} & \multicolumn{3}{c}{PaviaU} \\
                \cmidrule(lr){2-4} \cmidrule(lr){5-7} \cmidrule(lr){8-10}
                & OA($\%$) & NMI & $\mathcal{K}$ & OA($\%$) & NMI & $\mathcal{K}$ & OA($\%$) & NMI & $\mathcal{K}$ \\
                \midrule
                Baseline & 72.53 & 0.7538 & 0.6555 & 81.98 & 0.8607 & 0.7701 & 54.19 & 0.6340 & 0.4638 \\
                L & 72.47 & 0.7619 & 0.6545 & 81.85 & 0.8611 & 0.7651 & 60.13 & 0.6744 & 0.5282 \\
                MC         & 73.92 & 0.7843 & 0.6698 & 90.25 & 0.8898 & 0.8704 & 61.78 & 0.6743 & 0.5286 \\
                MC\&L      & \textbf{74.41} & \textbf{0.7902} & \textbf{0.6759} & \textbf{90.61} & \textbf{0.9101} & \textbf{0.8746} & \textbf{69.48} & \textbf{0.7490} & \textbf{0.6250} \\
                \bottomrule
               \multicolumn{10}{l}{\footnotesize Baseline is the original basis representation model proposed in~\cite{EDESC};}\\
               \multicolumn{10}{l}{\footnotesize L is the baseline model with local structure preservation;}\\
                \multicolumn{10}{l}{\footnotesize MC is the baseline model with non-local structure preservation;}\\
                \multicolumn{10}{l}{\footnotesize MC\&L is the baseline model with non-local\&local structure preservation constraints.}\\
            \end{tabular}%
        }
        \caption{Results of the ablation study}
        \label{tab:ablation_study}
    \end{table}

        The results from the ablation study indicate that applying only local structure preservation improves NMI across all three datasets. However, it slightly reduces accuracy on the Houston and Trento datasets, possibly due to the occasional smoothness in the original predictions. In contrast, applying only non-local structure preservation enhances performance across all datasets and metrics. Combining local and non-local structure preservation further boosts performance, as local structure preservation helps propagate the benefits of non-local preservation. Considering both structures allows for more effective learning of HSI structure compared to using either one alone.

    \subsubsection{Impact of the number of FINCH iterations}
\begin{figure}[ht]
        \centering
        \begin{subfigure}{0.48\textwidth}
         \centering
            \includegraphics[width=0.7\textwidth]{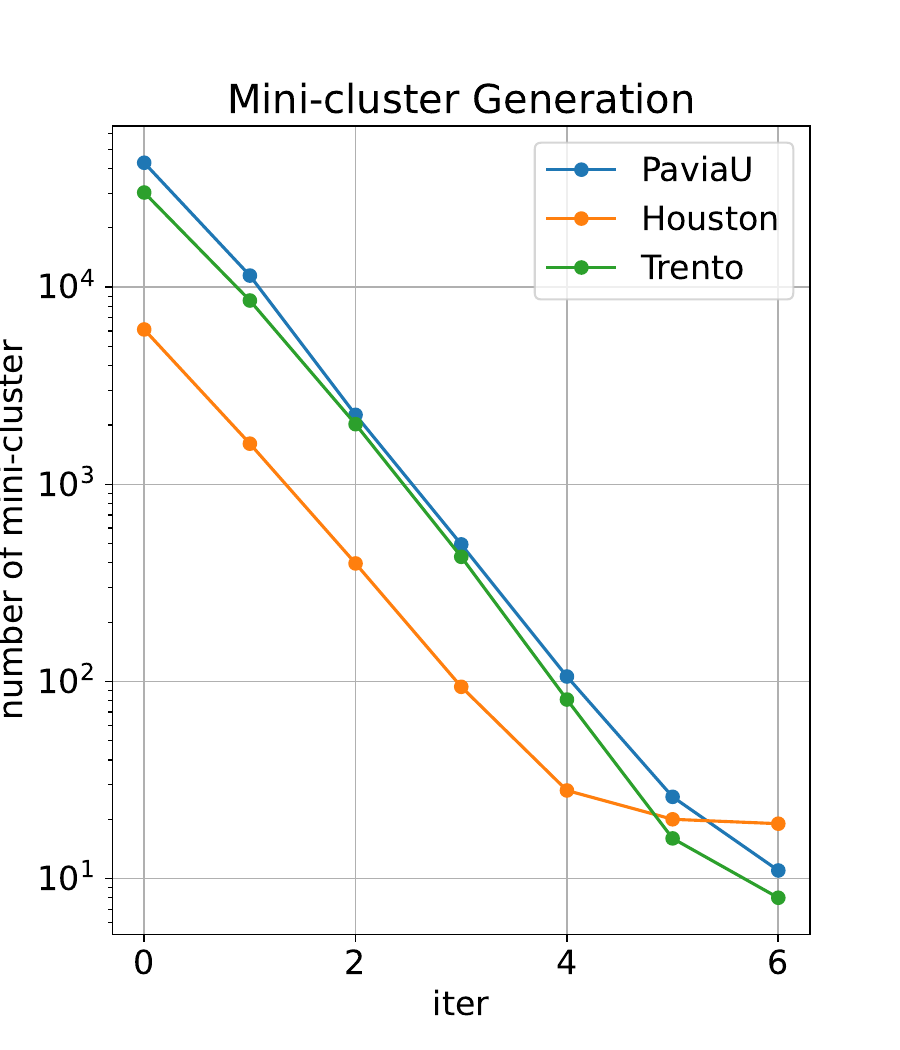}  
            \caption{Mini-cluster number by number of FINCH iterations.}
            \label{Fig:mini_cluster}
        \end{subfigure}
        \begin{subfigure}{0.48\textwidth}
         \centering
            \includegraphics[width=0.7\textwidth]{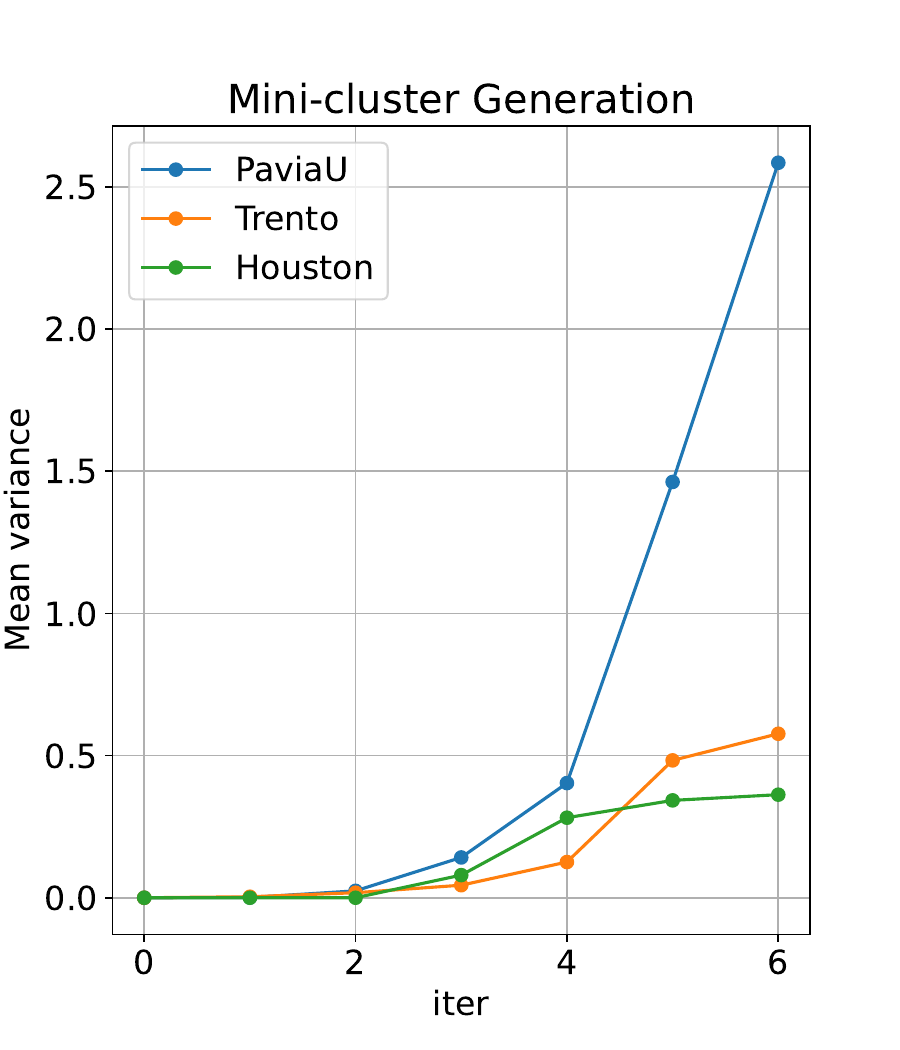}  
            \caption{Mini-cluster variance by number of FINCH iterations.}
            \label{Fig:mean_variance}
        \end{subfigure}
        \label{mini_cluster}
    \end{figure}

\begin{figure}[ht]
    \begin{subfigure}{0.325\textwidth}
     \centering
    \includegraphics[width=\textwidth]{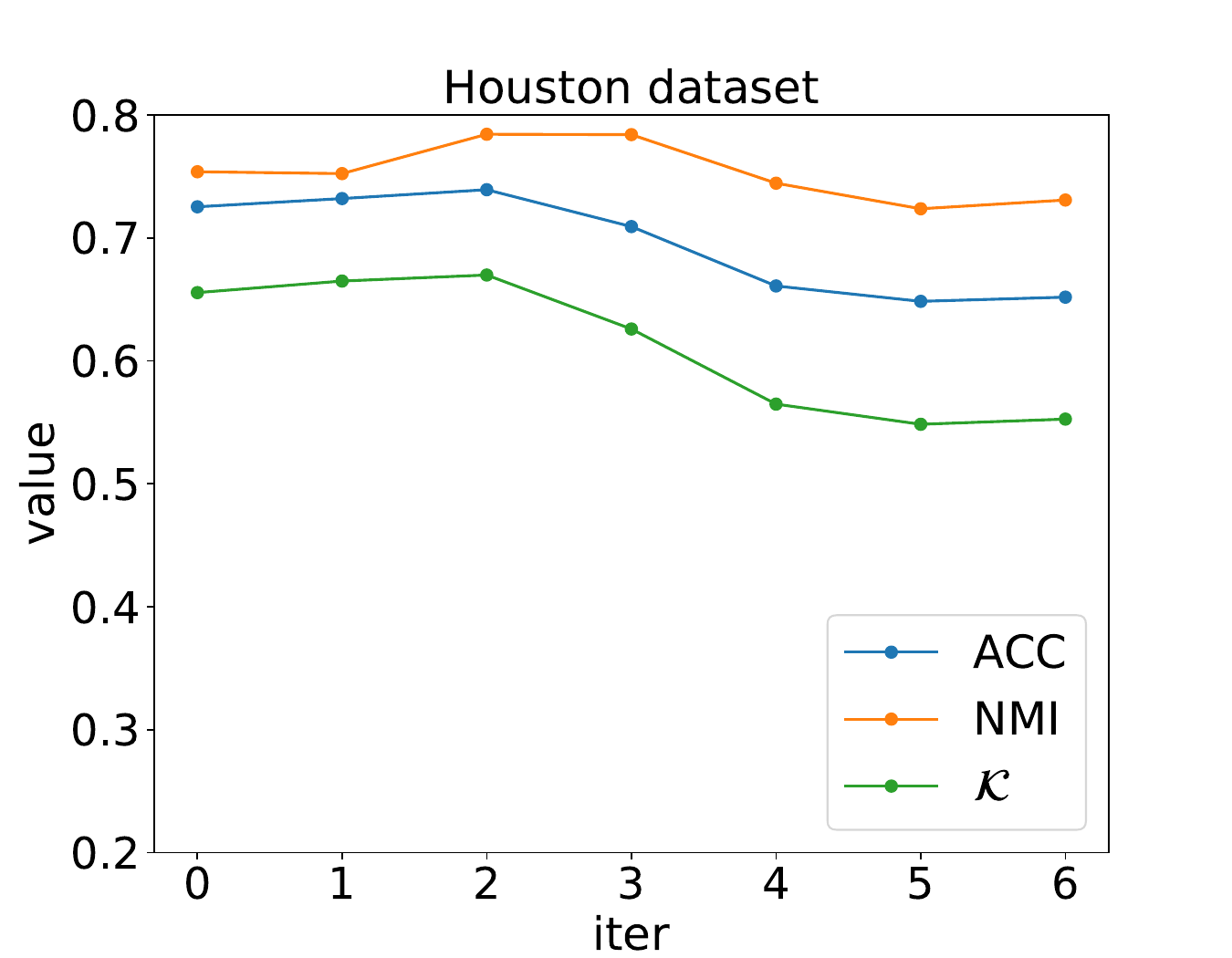}
    \end{subfigure}
    \begin{subfigure}{0.325\textwidth}
     \centering
    \includegraphics[width=\textwidth]{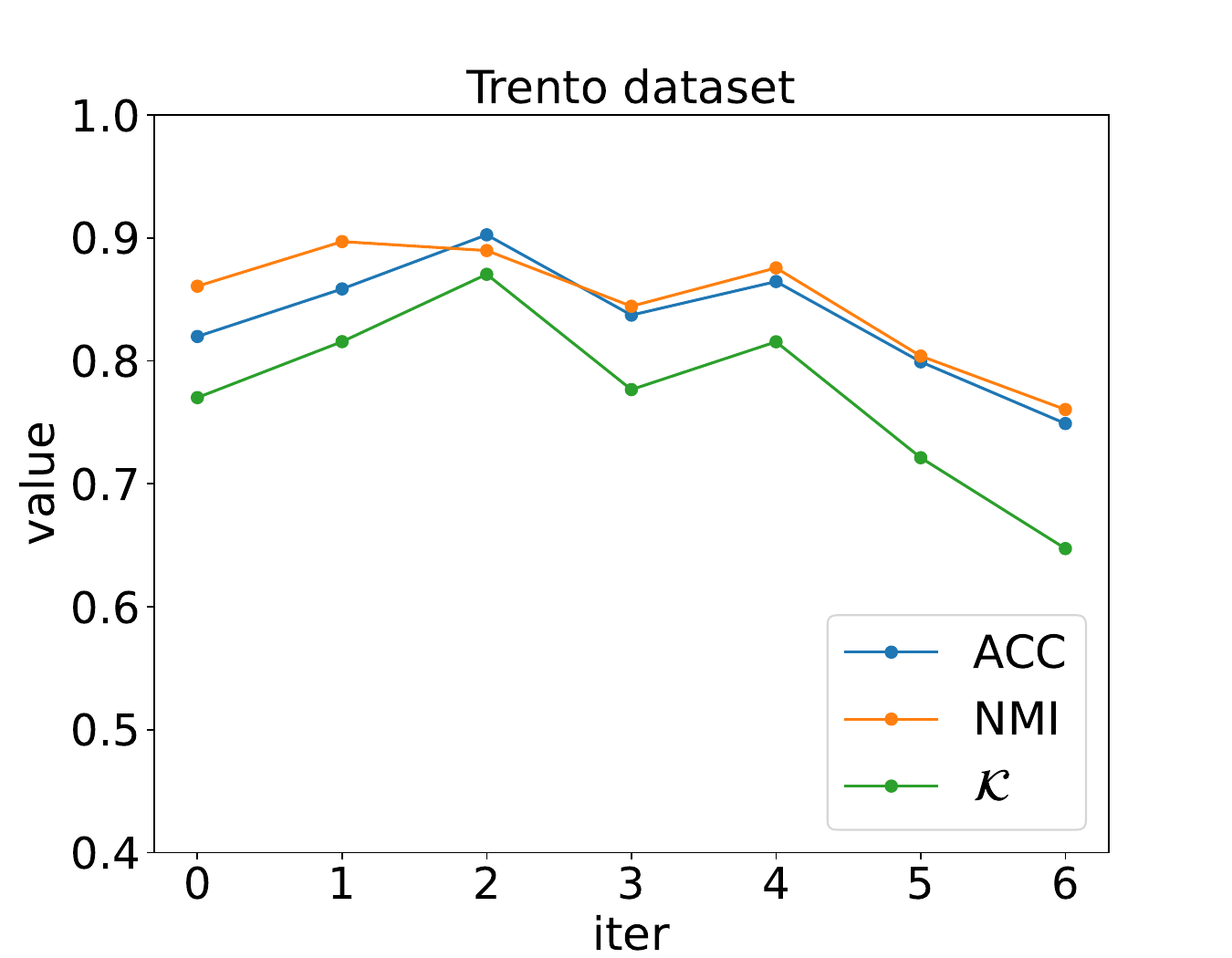}
    \end{subfigure}
    \begin{subfigure}{0.325\textwidth}
     \centering
    \includegraphics[width=\textwidth]{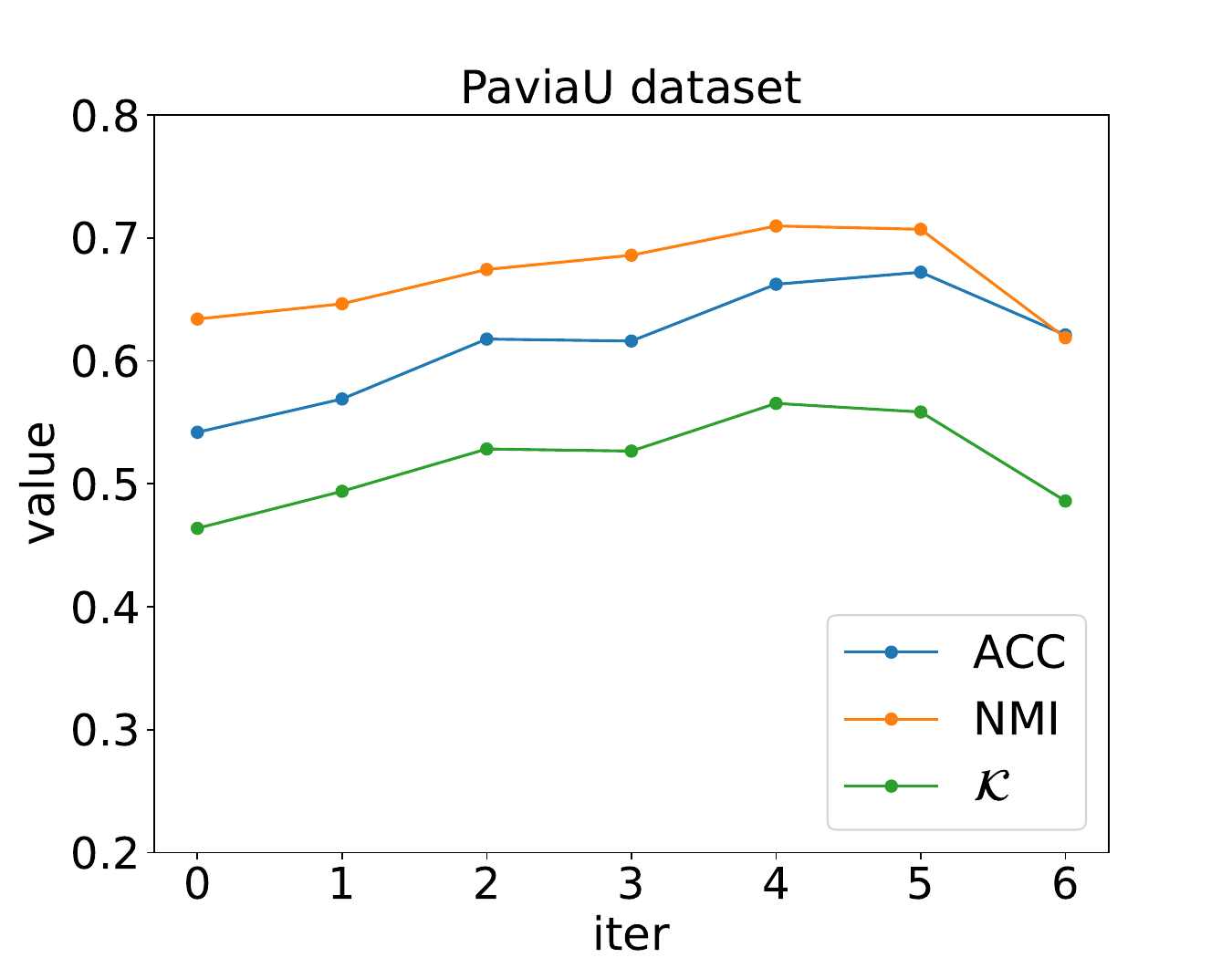}
    \end{subfigure}
    \caption{Performance with mini-clusters generated by number of FINCH iterations.}
    \label{Fig:mini_cluster_iteration_performance}
\end{figure}

    In our mini-cluster updating part, the algorithm FINCH~\cite{finch} is applied for mini-cluster generation. The number of mini-cluster decreases with the increase of FINCH iterations as shown in Fig.~\ref{Fig:mini_cluster}. Meanwhile, the error within each mini-cluster increases with the number of FINCH iterations. The mean-variance within every mini-cluster is calculated as shown in Fig.~\ref{Fig:mean_variance}. We observe that from the second iteration, the mean-variance starts to increase, indicating that there are more outliers within the mini-clusters. Specifically, the clustering performance of mini-clusters generated from different iterations is shown in Fig.~\ref{Fig:mini_cluster_iteration_performance}. We observe that the accuracy increases at first and then decreases. Specifically, for the Houston and Trento datasets we have the best performance after 2 iterations and then it decreases, which corroborates the findings related to variance. In conclusion, to ensure the quality of the mini-cluster and enhance clustering performance, the mini-clusters generated after the second iteration are applied.

\subsubsection{Impact of mini-cluster updating}
\begin{figure}[t]
    \begin{subfigure}{0.32\textwidth}
     \centering
        \includegraphics[width=\textwidth]{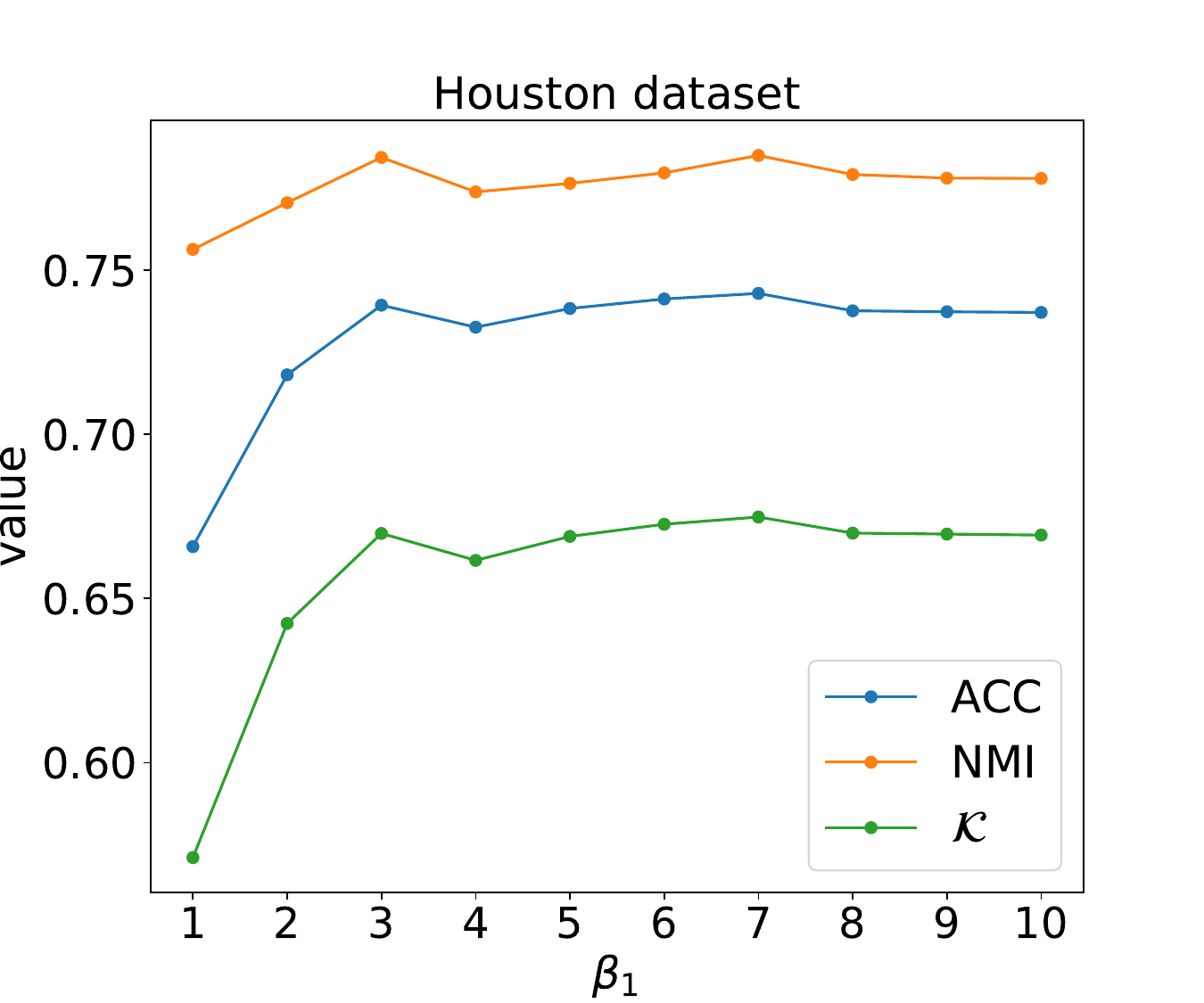}  
        \label{Fig:alpha_houston}
    \end{subfigure}
    \hfill  
    \begin{subfigure}{0.32\textwidth}
     \centering
        \includegraphics[width=\textwidth]{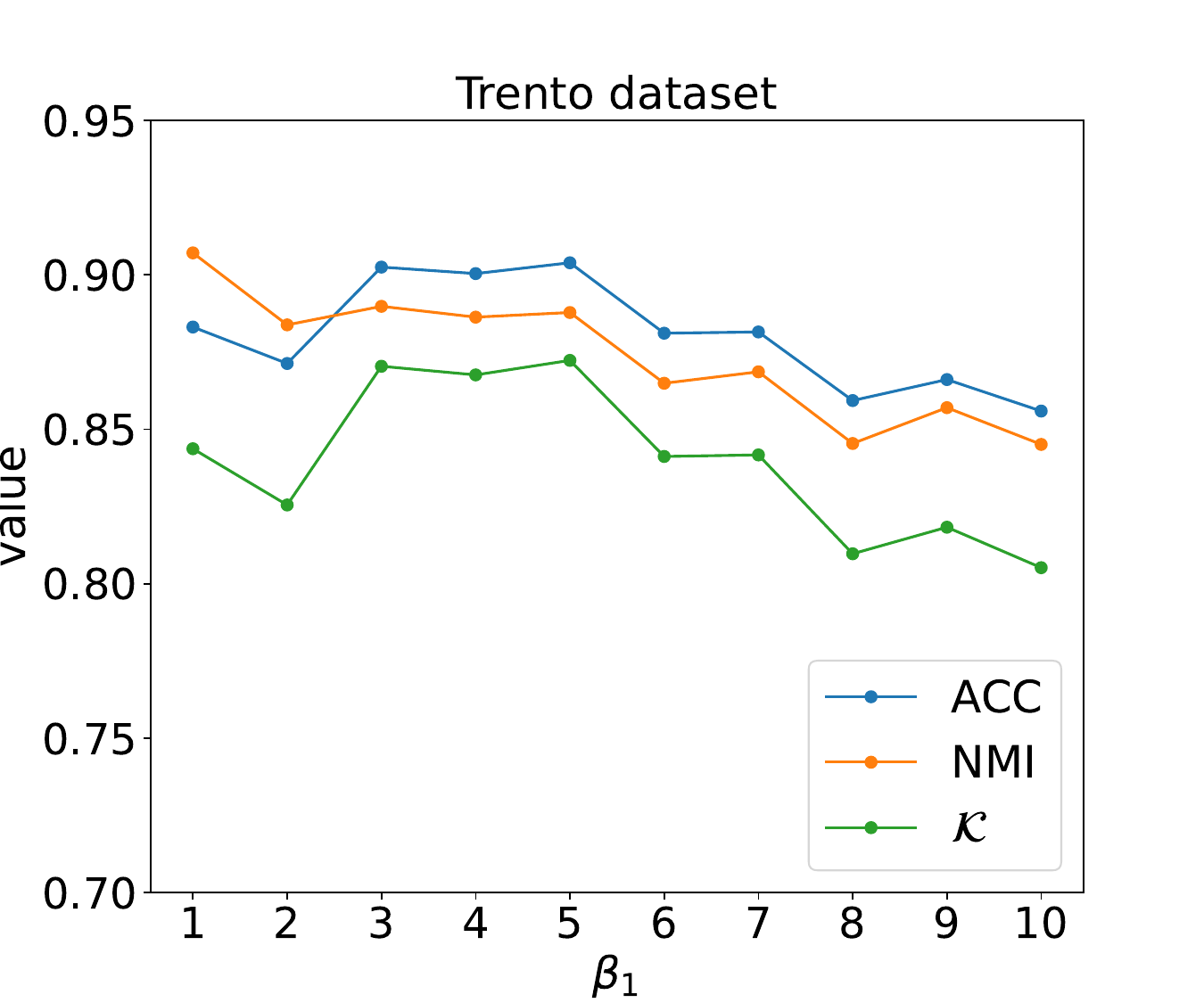}  
        \label{Fig:alpha_Trento}
    \end{subfigure}
        \hfill  
    \begin{subfigure}{0.32\textwidth}
     \centering
        \includegraphics[width=\textwidth]{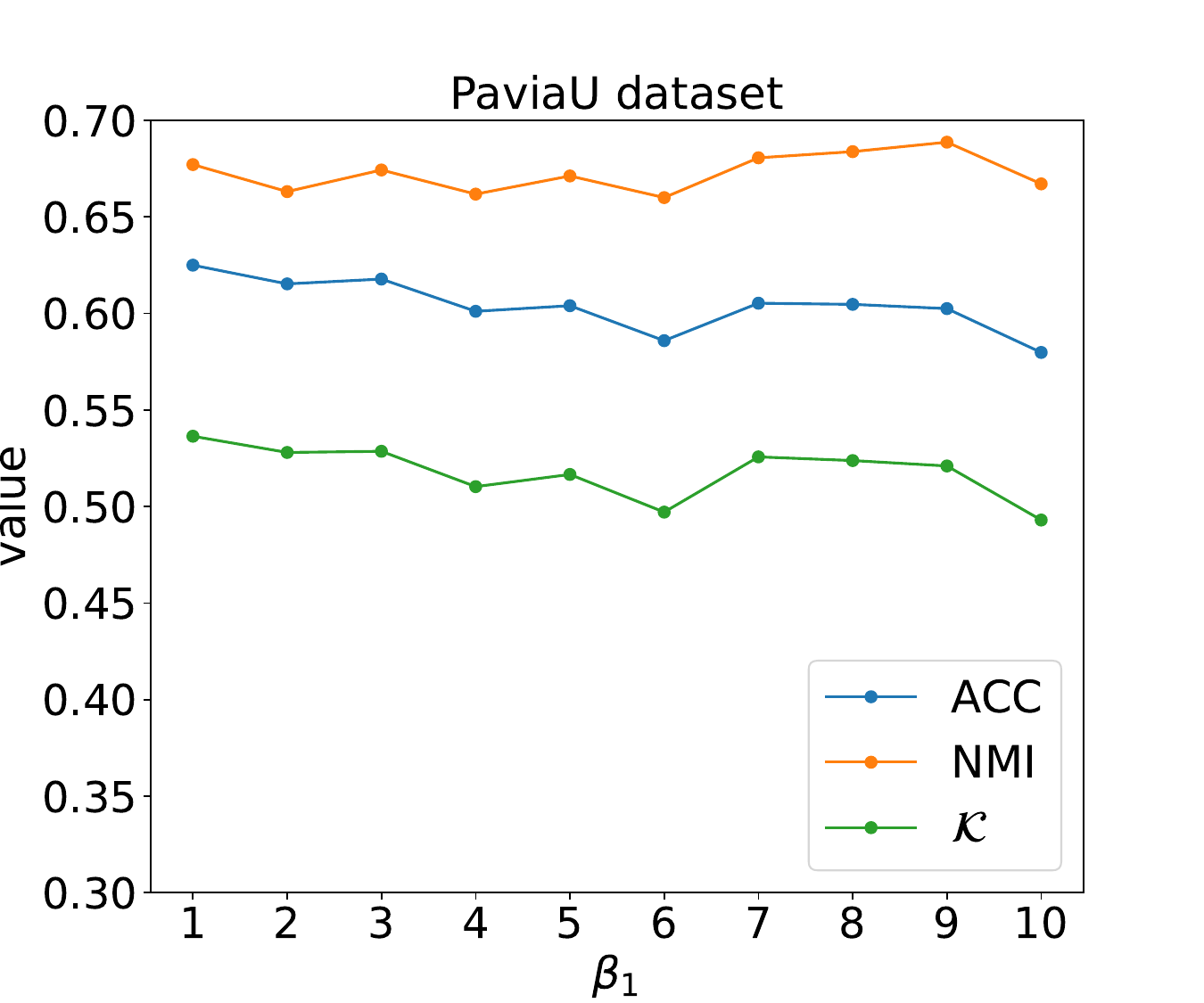}  
        \label{Fig:alpha_Pavia}
    \end{subfigure}
    \caption{The impact of mini cluster updating in clustering results.}
    \label{alpha_impact}
\end{figure}

The parameter $\beta_1$ controls the updating of the mini-cluster soft assignment which is important in the optimization. Here we compared the performance with different $\beta_1$ values on various datasets, the clustering result is shown in Fig.~\ref{alpha_impact}.
From the graph, it is evident that both too large or too small values of $\beta_1$ adversely affect clustering performance. For example, a small $\beta_1$ will decrease the performance on the Houston dataset. Meanwhile, a large $\beta_1$ leads to a lower accuracy on the Trento dataset. Since different datasets exhibit varying sensitivities to $\beta_1$, it is advisable to choose $\beta_1$ values between 3 and 5. In our experiments, we set $\beta_1$ to 3 across all datasets.

\subsubsection{Impact of local structure preservation}
\begin{figure}[t]
    \centering
    \begin{subfigure}{0.325\textwidth}
        \centering
        \includegraphics[width=\textwidth]{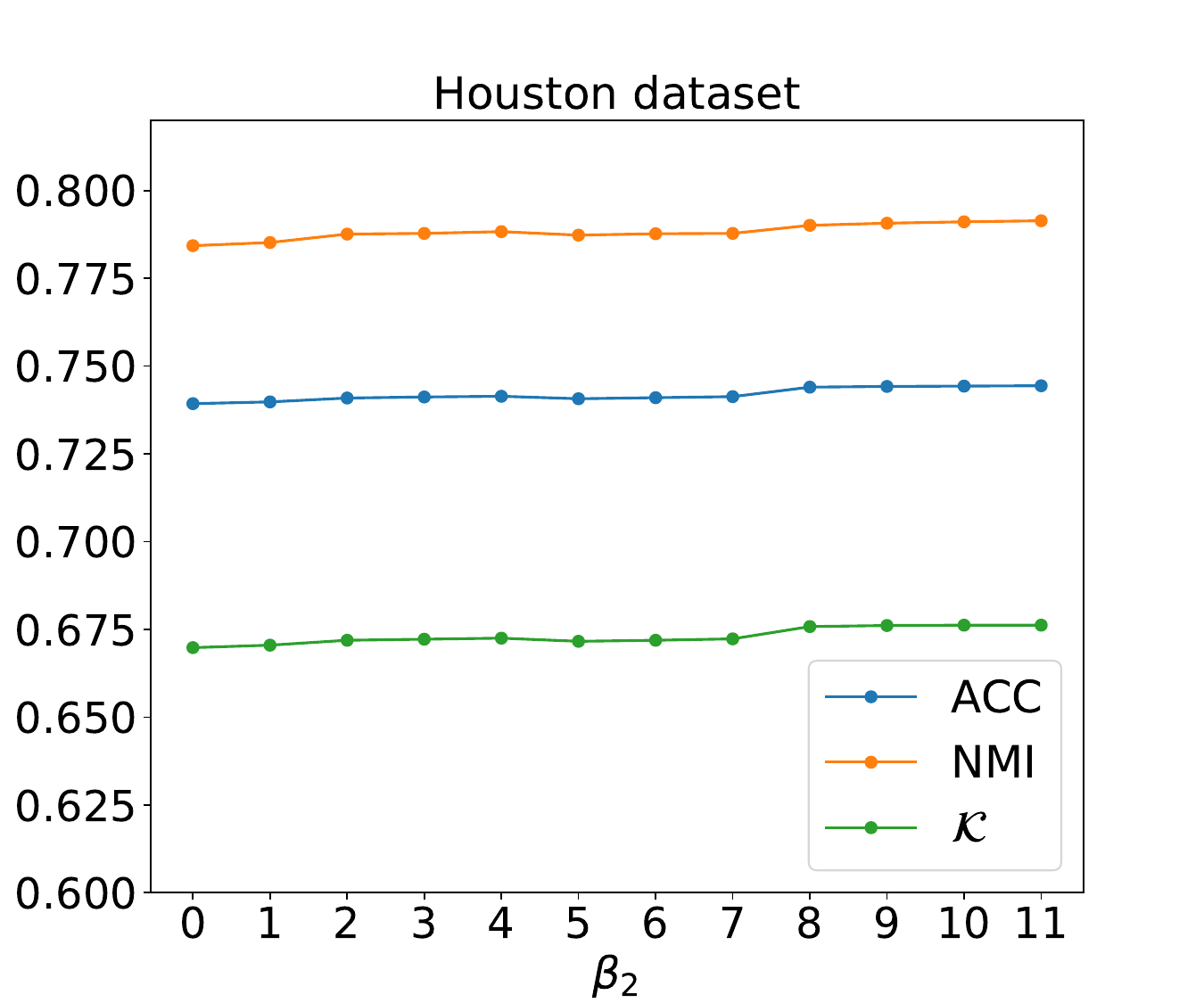}
        \label{Fig:houston-gamma}
    \end{subfigure}
    \hfill
    \begin{subfigure}{0.325\textwidth}
        \centering
        \includegraphics[width=\textwidth]{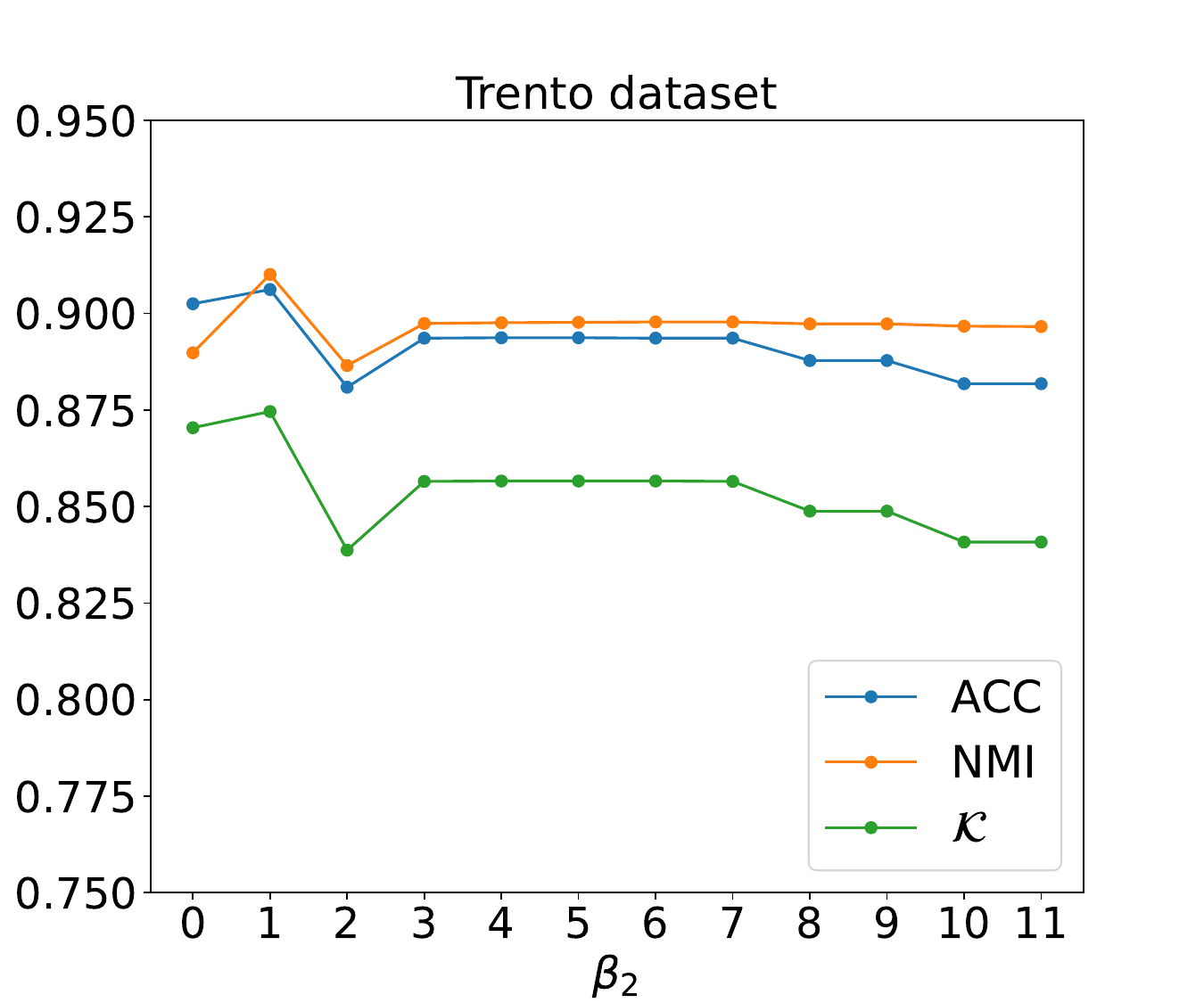}
        \label{Fig:trento-gamma1}
    \end{subfigure}
     \hfill
    \begin{subfigure}{0.325\textwidth}
        \centering
        \includegraphics[width=\textwidth]{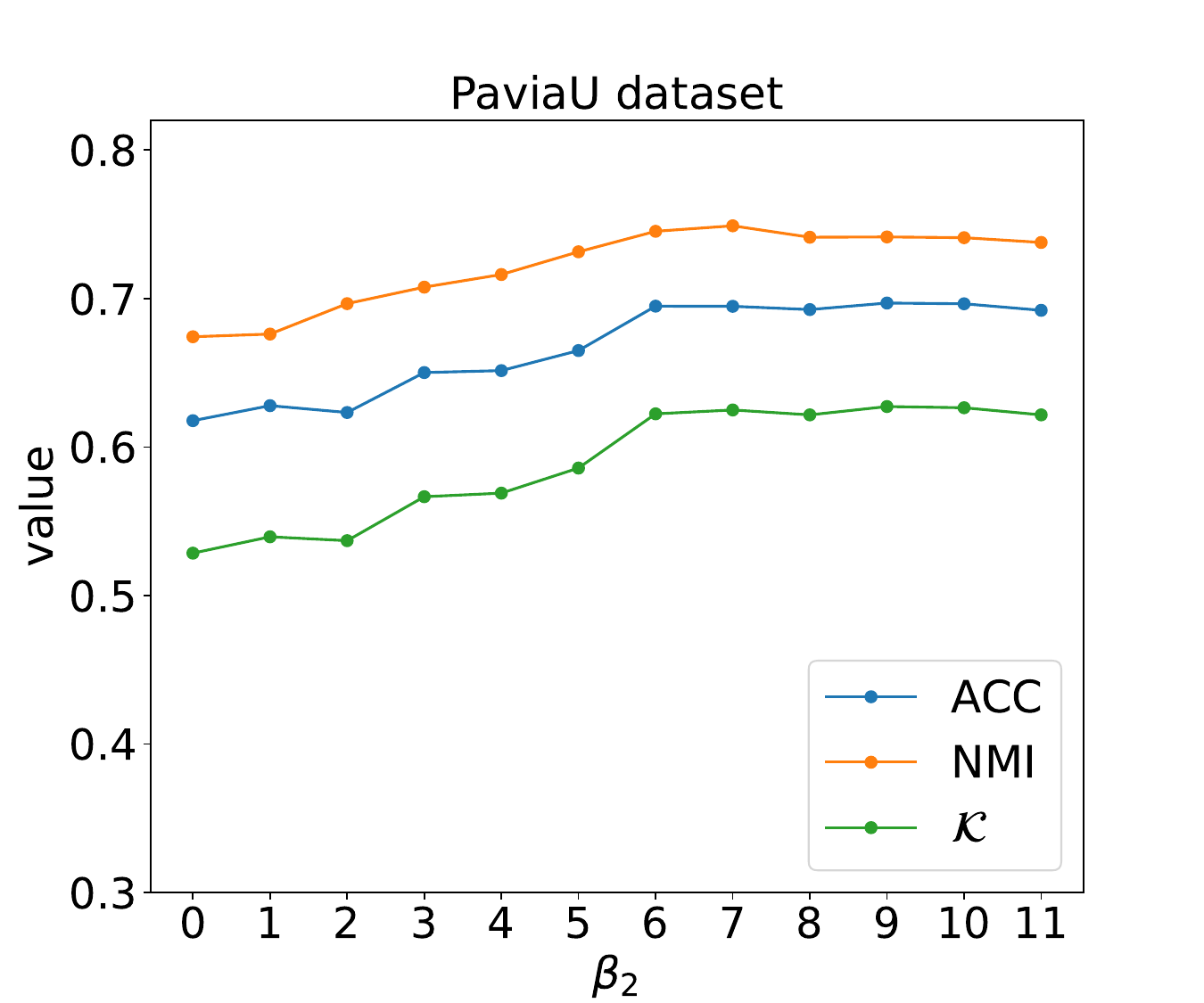}
        \label{Fig:trento-gamma2}
    \end{subfigure} 
    \caption{The impact of local structure preservation in clustering results.}
    \label{Fig:beta-impact}
\end{figure}

To accommodate the diverse inner local structures of different datasets, distinct settings are required for the local structure preservation module. To evaluate the influence of this module, we integrate it into a mini-cluster optimization model with varying weights. The clustering results with different local structure weights are shown in Fig.~\ref{Fig:beta-impact}.

The results show that the local structure preservation module effectively enhances clustering performance. Additionally, the optimal weight parameters vary across different datasets. For example, with simpler datasets like Trento, a small weight is adequate to achieve smooth clustering. On the other hand, for more complex datasets like Houston or PaviaU, a larger weight may be needed, typically within the range of [6, 10].

\subsubsection{Visualization of embedded representation}
\begin{figure}[ht]
    \centering
    \includegraphics[width=\linewidth]{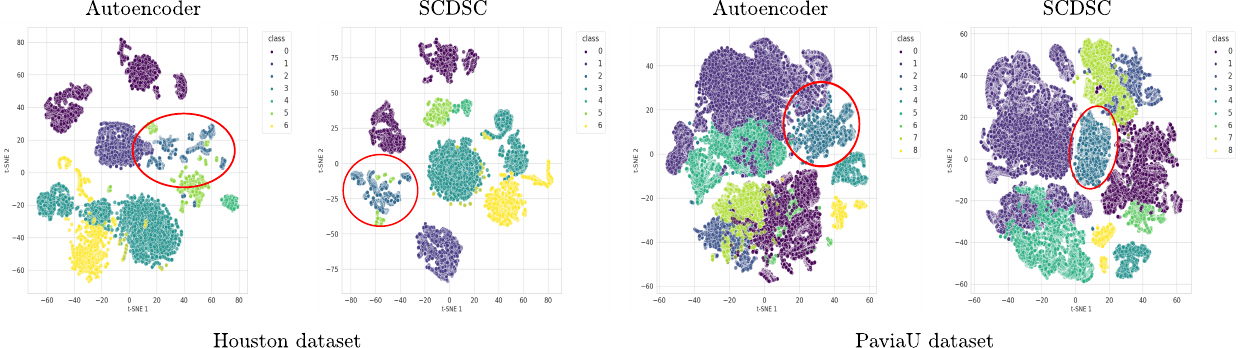}
    \caption{Visualization of the latent representation with t-SNE on Houston and PaviaU dataset.}
    \label{Visualization-representation}
\end{figure}

The t-SNE~\cite{van2008visualizing} visualization of the latent representation for the Houston and PaviaU datasets is shown in Fig.~\ref{Visualization-representation}. We compare the original latent representation produced by the autoencoder with the representation obtained by our proposed method. For the Houston dataset, the t-SNE visualization of the autoencoder's latent representation reveals significant overlap between some classes, resulting in unclear decision boundaries. Additionally, the intra-class points are loosely distributed, lacking tight clustering, which decreases the overall discriminative power. Our proposed method enhances the representation quality, reducing class overlap and improving intra-class compactness. For example, class 2 is well separated as highlighted in the red circle, leading to a more distinct and organized representation. Meanwhile, the latent representation for the PaviaU dataset from the autoencoder shows unclear decision boundaries, with many classes mixed. In contrast, our method generates a more discriminative representation, reducing the mixing of different classes, as illustrated by the red circle, ultimately leading to better class separation.

\subsubsection{Convergence analysis}
\begin{figure}[ht]
            \centering
            \begin{subfigure}{0.325\textwidth}
                \centering
                \includegraphics[width=\textwidth]{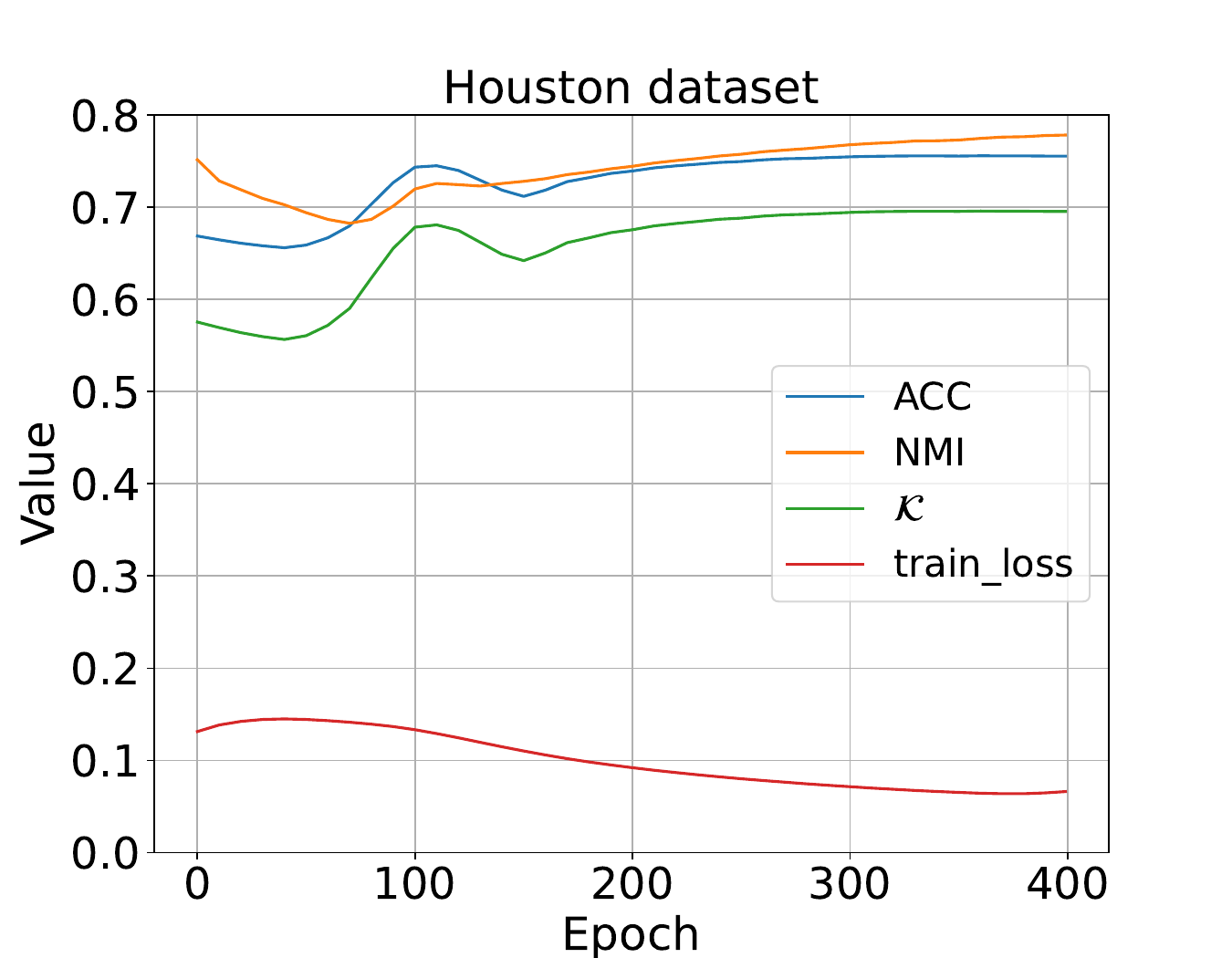}
            \end{subfigure}
             \hfill
            \begin{subfigure}{0.325\textwidth}
                \centering
                \includegraphics[width=\textwidth]{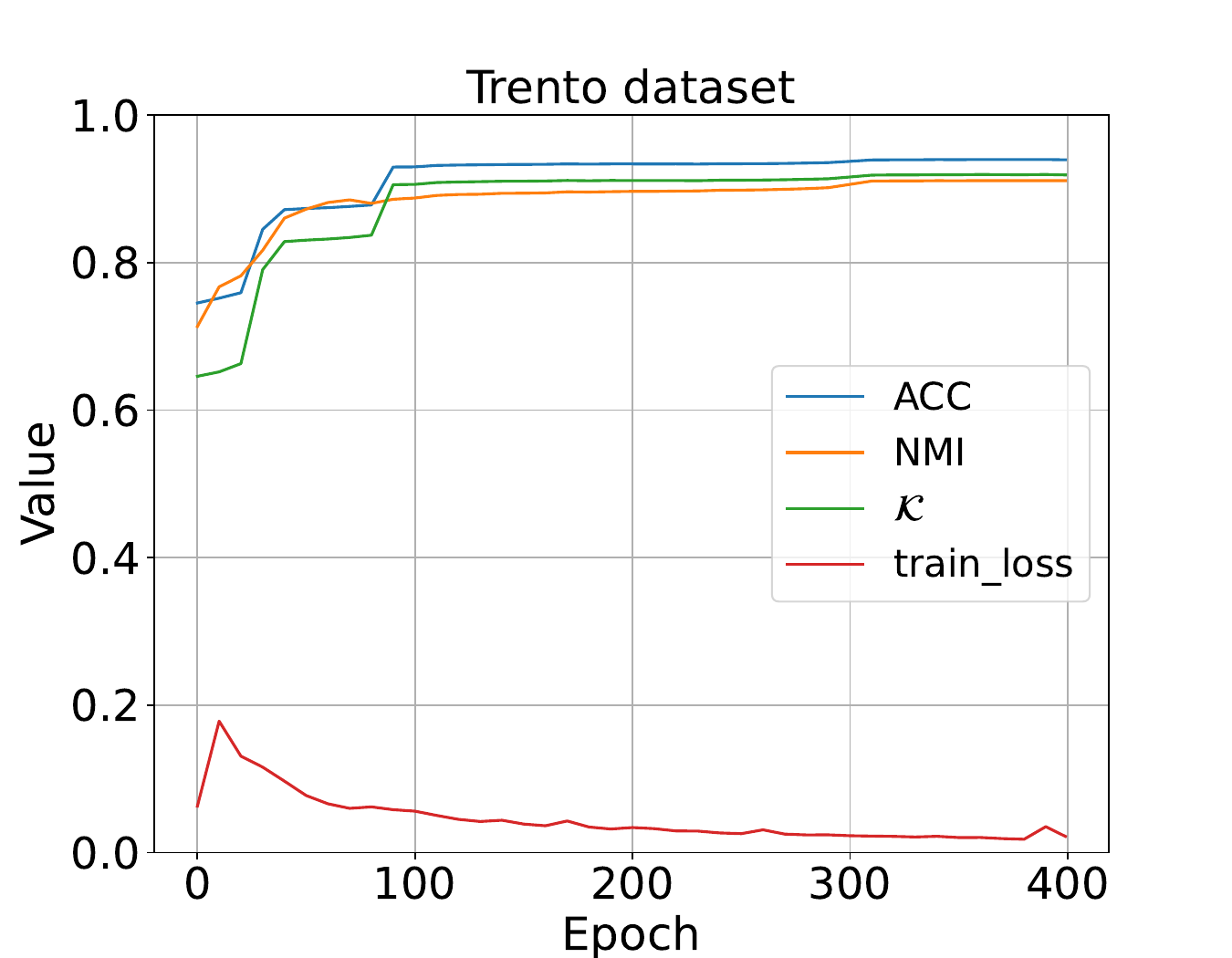}
            \end{subfigure}
             \hfill
            \begin{subfigure}{0.325\textwidth}
                \centering
                \includegraphics[width=\textwidth]{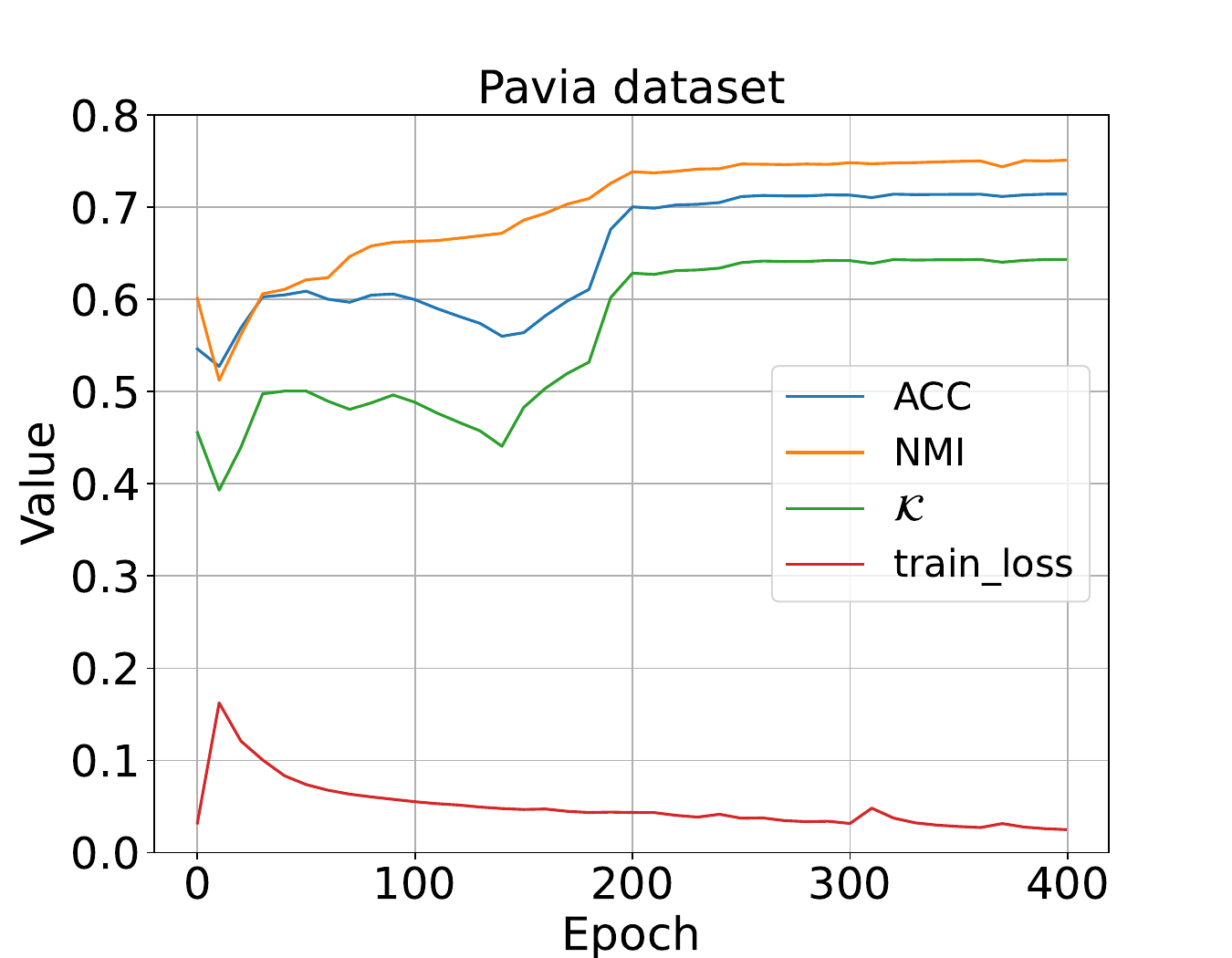}
            \end{subfigure} 
            \caption{Training Loss and Accuracy Curve.}
            \label{Fig:convergence}
        \end{figure}

    To validate our model's convergence, we plot the training curve on the test datasets. The results are shown in Fig.~\ref{Fig:convergence}. From the training curve, we observe that as the training iterations increase, the training loss decreases while the training accuracy increases, eventually plateauing. This indicates that our model converges well.

\section{Conclusion}
    In this paper, we present a concise review of model-based and deep clustering methods, including both purely data-driven and model-aware approaches. 
    Our primary contribution is a scalable context-preserving model-aware deep clustering approach for hyperspectral images. 
    The proposed method learns the subspace basis under the supervision of both local and non-local structures inherent to hyperspectral image data, allowing these structures to mutually reinforce each other during training. 
    Our approach achieves clustering with a computational complexity of $\mathcal{O}(n)$, making it scalable for large-scale data. 
    As opposed to the previous state of the art, in our method both the local and non-local structure preservation constraints optimize the entire clustering process in an end-to-end manner and provide stronger guidance for model optimization. 
    Experimental results on three benchmark hyperspectral datasets demonstrate that our method outperforms state-of-the-art approaches in terms of clustering performance.
\bibliographystyle{elsarticle-num} 
\bibliography{reference}

\end{document}